\documentclass[conference]{IEEEtran}
\IEEEoverridecommandlockouts
\usepackage{cite}
\usepackage{amsmath,amssymb,amsfonts}
\usepackage{algorithmic}
\usepackage{graphicx}
\usepackage{textcomp}
\usepackage{xcolor}

\ifCLASSOPTIONcompsoc
    \usepackage[caption=false, font=normalsize, labelfont=sf, textfont=sf]{subfig}
\else
\usepackage[caption=false, font=footnotesize]{subfig}
\fi
\usepackage[colorlinks=true,allcolors={blue}]{hyperref}

\def\BibTeX{{\rm B\kern-.05em{\sc i\kern-.025em b}\kern-.08em
    T\kern-.1667em\lower.7ex\hbox{E}\kern-.125emX}}
\begin{document}

\title{Comparative Study of Q-Learning and  NeuroEvolution of Augmenting Topologies for Self Driving Agents
}

\author{\IEEEauthorblockN{1\textsuperscript{st} Arhum Ishtiaq}
\IEEEauthorblockA{\textit{DSSE, Computer Science} \\
\textit{Habib University}\\
Karachi, Pakistan \\
ai05182@st.habib.edu.pk}
\and
\IEEEauthorblockN{2\textsuperscript{nd} Maheen Anees}
\IEEEauthorblockA{\textit{DSSE, Computer Science} \\
\textit{Habib University}\\
Karachi, Pakistan \\
ma05156@st.habib.edu.pk}
\and
\IEEEauthorblockN{3\textsuperscript{rd} Neha Jafry}
\IEEEauthorblockA{\textit{DSSE, Computer Science} \\
\textit{Habib University}\\
Karachi, Pakistan \\
nj05165@st.habib.edu.pk}
\and
\IEEEauthorblockN{4\textsuperscript{th} Sara Mahmood}
\IEEEauthorblockA{\textit{DSSE, Computer Science} \\
\textit{Habib University}\\
Karachi, Pakistan \\
sm05155@st.habib.edu.pk}
}

\maketitle




\section{Introduction}

Autonomous driving vehicles have been of keen interest ever since automation of various tasks started. Humans are prone to exhaustion and have a slow response time on the road, and on top of that driving is already quite a dangerous task with around 1.35 million road traffic incident deaths each year \cite{cdc_trfdata}. It is expected that autonomous driving can reduce the number of driving accidents around the world which is why this problem has been of keen interest for researchers.

Currently, self-driving vehicles use different algorithms for various sub-problems in making the vehicle autonomous. We will focus  reinforcement learning algorithms, more specifically Q-learning algorithms and NeuroEvolution of Augment Topologies (NEAT), a combination of evolutionary algorithms and artificial neural networks, to train a model agent to learn how to drive on a given path. This paper will focus on drawing a comparison between the two aforementioned algorithms. 

\section{Technical Background}\label{Bg}
\subsection{Reinforcement Learning}\label{rl}
\begin{figure}[ht]
    \centering
    \includegraphics[width=\linewidth]{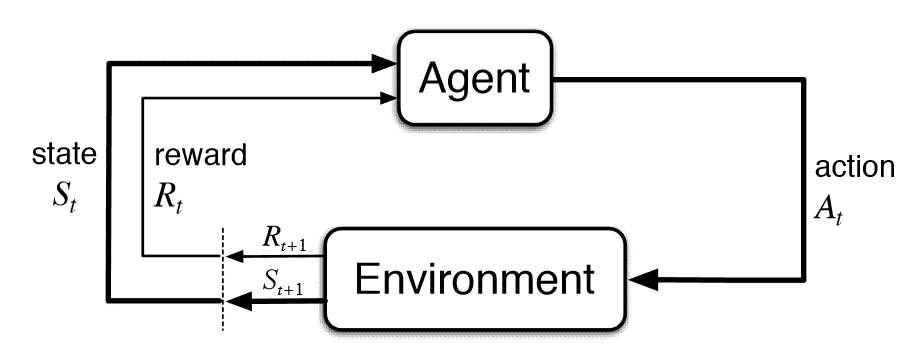}
    \caption{Q learning algorithm}
    \label{fig:qlearn}
\end{figure}

 Reinforcement Learning (RL) algorithms are unique and distinctive in the sense that they do not fall under the category of the traditional supervised or unsupervised learning. RL algorithms are not provided with any training data. Instead learning is based on a reward and punishment paradigm learned by interacting with their environment. An agent chooses an action and based on the action receives a reward (positive or negative). Then according to the learned values from rewards, the agent chooses the next action in a way that maximizes the cumulative reward. 
 A visual representation of how RL algorithms works is given in Fig. \ref{fig:qlearn}.

\subsection{Q-Learning}\label{q}
Q-Learning is a model-free reinforcement learning algorithm which involves direct learning of the Q-values. The algorithm is based on maximizing the total reward value, the Q-value, and the agent moves in such a way as to maximize the Q-value. The Q-value is updated using the following rule:
\begin{equation}\label{be}
    Q(s,a) = r(s,a)+\gamma max_a Q(s',a)
\end{equation}
In this equation, $s$ and $s'$ are the current state and the next state, respectively, reached by performing the action $a$. Further, $r(s,a)$ denotes the immediate reward received by performing action $a$, and $\gamma$ denotes the discounting factor which controls the effect of the reward in the distant future. The $max_a Q(s',a)$ term in the equation finds the maximum reward possible from the next state.

As the Q-value depends on the Q-value of the future state the Q-value of state $s$ comes out to be:
\begin{equation}
    Q(s,a) \xleftarrow{} \gamma Q(s',a)+\gamma^2 Q(s'',a)+\ldots \gamma^n Q(s''^{\ldots n},a)
\end{equation}
The value $\gamma \in [0,1]$, if $\gamma = 1$ then all future rewards have an contribute to the total reward and vice versa for $\gamma = 0$.

We start the learning process by using arbitrary values and converge to an optimum after learning from the negative and positive rewards.

\subsection{Neuro-evolution}\label{neuro}
Neuro-evolution, in technical terms, is defined as the evolution of ANNs using EA. Instead of relying on a fixed structure for a neural network, neuro-evolution basically borrows the idea of biological evolution from nature and combines it with the power of brain to provide solutions to problems, by recombination of individuals in a population, over several generations. This population based learning, through mutation and crossover then helps evolve population with fittest individuals, eventually leading to the solution where the best individual(highest fitness) is found that exhibits a desired behaviour on a given task \cite{Huang}.

\subsubsection{Evolutionary Algorithms}\label{EA}
\begin{figure}[htbp]
    \centering
    \includegraphics[width=0.45\textwidth]{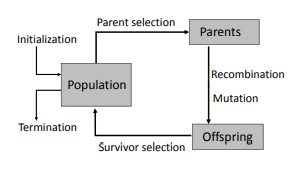}
    \caption{General framework of an evolutionary algorithm}
    \label{fig:2}
\end{figure}

EAs are optimization algorithms that imitate the natural process of evolution to find the optimal solution(s) to a given computational problem by maximizing or minimizing a  particular function. By maintaining a population of candidate solutions, EA borrows the idea from the process of reproduction and natural selection to find the `fittest' solution. The general framework of EA is shown in Fig. \ref{fig:2} and the algorithm proceeds as follows \cite{Huang}:
\begin{enumerate}
    \item \textit{Initialization:} A population of $n$ candidate solutions is initialized either randomly or using a heuristic that are valid for the given problem.
    \item \textit{Fitness Calculation:} Each individual/chromosome in the population is evaluated against a criteria and a measure of fitness is assigned to it.
    \item \textit{Parent Selection and Crossover:} Individuals are selected as parent chromosomes from the population using a selection criteria (for example, fitness proportionate, elitism etc.) which are then recombined to produce offspring. Invalid or damaged offspring are either discarded and recreated or prevented from being produced.
    \item \textit{Mutation:} Offspring are mutated by a mutation rate and then evaluated where fitness is calculated and assigned to each offspring.
    \item \textit{Survivor Selection:} Using a replacement criteria (for example: fitness based, elitism etc), offspring replace parents in the population.
    \item \textit{Loop:} Steps 2 to 5 are repeated until the termination criteria (solution found or maximum number of evaluations) is reached.
\end{enumerate}

\subsection{NeuroEvolution of Augmenting Topologies (NEAT)}\label{NEAT}
NEAT is an evolutionary algorithm that creates Artificial Neural Networks. NEAT is a complexification algorithm, which means that it starts out with very simple, minimal neural networks and over evolutionary generations, increases the number of neurons and connections between them. Complexification algorithms like NEAT are beneficial to use because they make it easier to evolve networks without changing their behaviour drastically and therefore avoiding large networks that are slower to adapt \cite{Huang}. 

In the current implementation of NEAT-Python, a population of individuals is maintained where each individual contains two lists of genes:
\begin{enumerate}
    \item \textit{Node genes}, each of which identify a single neuron.
    \item \textit{Connection genes}, each of which refers to a single connection between two neurons, specifying where a connection comes into and out of, the weight of such connection, whether or not the connection is enabled, and an innovation number which allows for finding the corresponding genes during crossover \cite{Stanley}.
\end{enumerate}

To evolve a solution to a problem using NEAT, the fitness function must be provided by the user which computes a single real number indicating how fit an individual genome is. The algorithm progresses through number of generations specified by the user, with each generation being produced by reproduction and mutation of the most fit individuals of the previous generation, just like in standard evolutionary algorithms \cite{Stanley}. Details of the evolutionary operators with respect to NEAT will be discussed in section \ref{algo_neat}.

\section{Related Work}\label{PrevWork}
\subsection{Q-Learning}

A lot of work regarding reinforcement learning for autonomous driving vehicles has been done in the past. In paper \cite{kiran}, Kiran et. al have discussed various reinforcement learning approaches being used for self-driving agents, including Q-learning . Q-learning is based on learning values for state-action pairs, stored in tables, in order to converge to an optimal solution.

In paper \cite{vitelli}, Vitelli and Nayebi compared the discrete Q-learning approach for self-driving agents to the Deep Q-learning approach, an extension of q-learning which uses a CNN to estimate q-values. Their best CNN based Deep Q-Network was seven layers deep but still was unable to train the agent enough to make a complete turn. On the other hand, their discrete Q-learning algorithm did achieve convergence. Moreover, the discrete Q-learning algorithm despite having a much lower average reward, achieved much higher average speeds and the maximum speed was comparable with the DQN algorithm.

\subsection{NEAT}
There has been significant amount of work done on evolving the driving behaviour of vehicles using neuro-evolution techniques. In \cite{Togelius}, ANNs were used for evolving the behavior by using a 3-layered neural network with fixed topology. 

Ebner and Tiede in \cite{Ebner} used Genetic Programming, which is another EA, to train a single vehicle around the track as fast as possible while avoiding the obstacles. 

In \cite{Drchal}, HyperNEAT, a variant of NEAT utilizing indirect encoding, is used as the EA to evolve network weights as well as the topologies.

In \cite{Talamini}, the ANN evolution scheme using NEAT is being employed and its impact is compared by using global fitness as well as local fitness in terms of the speed, safety and efficiency of the vehicle. 

\section{Implementation}\label{algo}

\subsection{Q-Learning}
We implemented Q-learning algorithm described in section \ref{q} on the self driving car agent problem. Here we will discuss how we formulated this problem to maximize the score and distance covered by the car.

\subsubsection{\textbf{Action}}
For the scope of this project our agent could take six possible actions:
\begin{itemize}
    \item Speed up: If the  agent chooses this action, the speed of the car is incremented by 2 units and this speed is also considered as the distance covered by the agent in that particular action. The direction of the car is maintained.
    \item Turn Left: If the agent chooses this action then the car turns by an angle of 15 degrees to the left without covering any distance.
    \item Turn Right: If the agent chooses this action then the car turns by an angle of 15 degrees to the right without covering any distance.
    \item Slow down: If the speed of the car exceeds 10 then this action reduces the speed of the car by 2 units.
    \item Turn Left and speed up: If the agent chooses this action then the car turns by an angle of 15 degrees to the left and the speed of the car is incremented by 2 units, meaning distance of 2 units is also covered
    \item Turn Right: If the agent chooses this action then the car turns by an angle of 15 degrees to the right and the speed of the car is incremented by 2 units, meaning distance of 2 units is also covered
\end{itemize}

\subsubsection{\textbf{State}}
Q learning requires states of the agent to be discrete, where as in most of the real world scenarios like this, the state of the agent consists of continuous values. In our experiment we have 5 continuous space states obtained from the five radar sensors but we can't use these continuous values to make the state table. To overcome this we discretized our observed states into buckets. We made 11 buckets for each value of the radar. The dimensions of the state table now depends on the number of buckets created for the space states and number of actions, therefore the final dimensions of our state table are $(11 ,11,11,11,11,3)$

\subsubsection{\textbf{Punishment}}\label{pu}
The major aim of the agent is to drive on the path for maximum distance without colliding with the edges of the track. The agent is penalized with a score of negative 1000 plus the distance it has covered on the path divided by 10. Throughout the algorithm, the agent tries to maximize the score, thus learns to cover the maximum distance while staying on path.

\subsubsection{\textbf{Reward}}\label{re}
\begin{figure}[ht]
    \centering
    \includegraphics[width=0.45\textwidth]{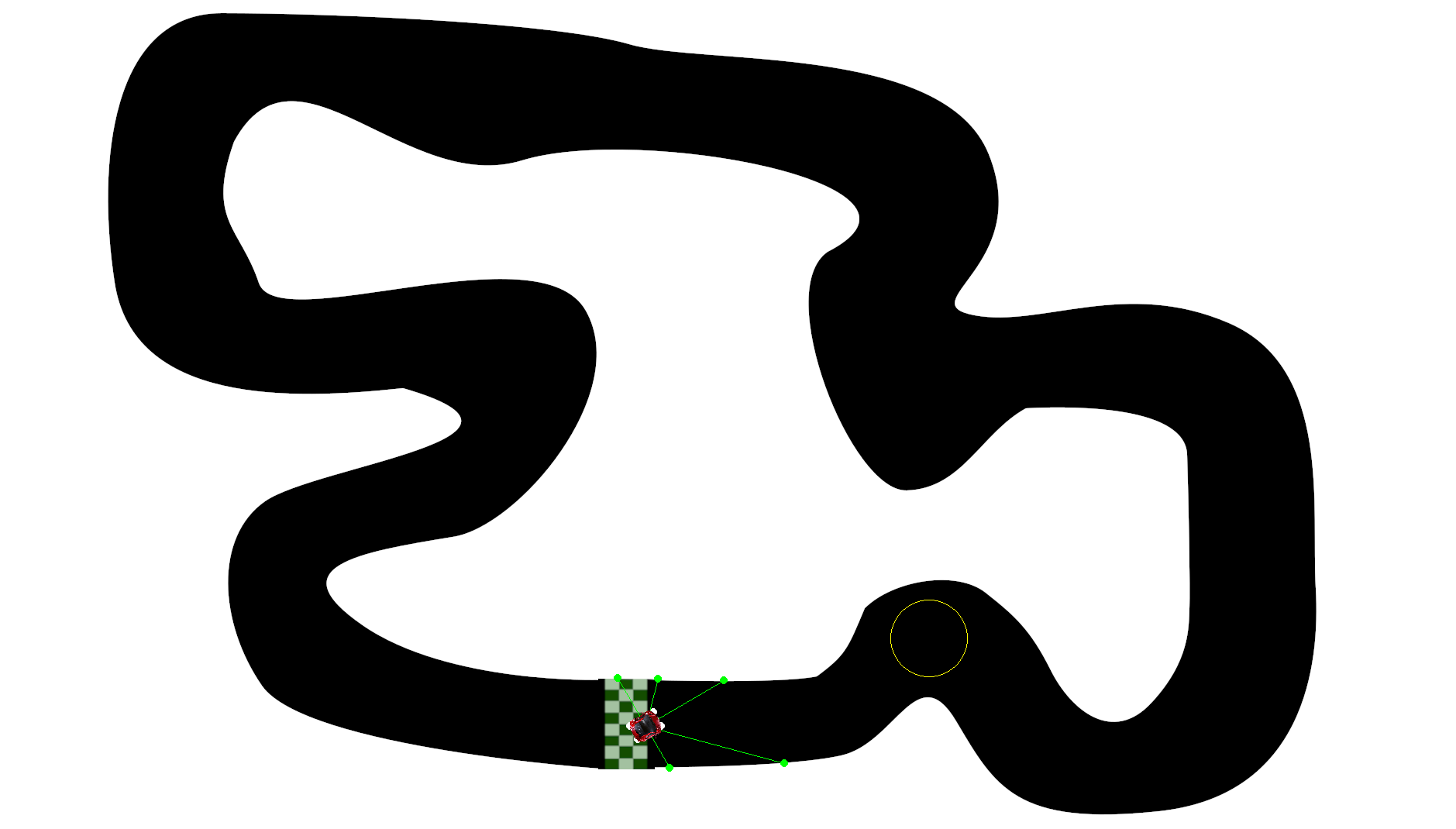}
    \caption{A checkpoint on map 2}
    \label{fig:check}
\end{figure}

The agent is rewarded with positive 10 points once it crosses the a checkpoint, shown as a circle on the track in figure \ref{fig:check}. When the car crosses the finishing line on the simulator an additional 50 points is added to the reward. Apart from this, increasing the distance reduces the penalty for colliding with the edges of the track. 

Initially our agent was rewarded to survive in the environment without colliding with the edges of the path. The problem we faced with this was that our agent learnt to increase the score by just staying in it's place or moving in circles around it's start position. To overcome this we changed our reward and punishment methodology to the one described in \ref{re} and \ref{pu}. This approach provides an incentive to the agent to maximize the distance covered and reach to the last checkpoint.

For each state, the agent chooses the next action based on the q-values of all the legal actions. The q-values are calculated using eq. \ref{be}. To introduce randomness in the initial stages of training we implemented an epsilon greedy technique where a parameter epsilon was introduced which signifies the percentage of random actions chosen by the agent. The epsilon decreases over the course of training and is left at 0.01 which means the final model will take random actions $1\%$ of the time.

\subsection{NEAT}\label{algo_neat}
\subsubsection{Chromosome Representation}
Our phenotype is a fully trained neural network in which the total number of neurons will be somewhere close to a sum of the $num\_hidden$, $num\_input$ and $num\_output$ configuration parameters, representing the neurons in the hidden, input and output layers of our network, respectively. The genotype will be a combination of two unique gene sets comprising of:
\begin{itemize}
    \item \textit{Nodes} - analogous to individual neurons, these nodes will be the main building blocks that will be connected to make a complete neural network
    \item \textit{Edge connections} - the connections between arbitrary nodes which allow the mapping of data from one layer to the next in the neural network. 
\end{itemize}
Since the goal of this implementation is to find the best model that will be able to fully loop a car sprite around any given track, the best model genotype will then be subsequently given control of our car instances and allowed to control its movement and ensure that it loops around the track using the radar-like ``sensors" of the car  that we implemented to give some level of awareness to our model in terms of what is around it. 

\subsubsection{Fitness Function}
Our fitness function has to reward the ability of our car to go farther and do that faster. Therefore, after testing multiple iterations of our fitness function, we decided that a scaled product of distance covered and car speed would most effectively allow NEAT to not just  improve the distance covered but also decrease the time taken to do so, as the generations progress through the learning process. 
\begin{equation} \label{eq:2}
    fitness() = distance_i\times speed_i\times 10^{-6}\\
\end{equation}

In (\ref{eq:2}), the $i$ represents the index of the population which goes from $0$ till $size(population)-1$. 

\subsubsection{Crossover}
While crossover in usual evolutionary approaches is straightforward, in the case of NEAT, we have to perform crossover between two networks that may have wildly differing structures. Therefore, to help solve for this, NEAT assigns a global key to each gene after any structural mutation (addition or removal of either a node or edge connection) occurs which is then used as a historical marker of sorts in order to ensure compatibility between two genes when the crossover part happens.

\subsubsection{Mutation}
We mutate our chromosome representation in two main aspects that allow for a stochastic increase in exploration due to the new sample space being created as a result of the following types mutations:
\begin{itemize}
    \item \textit{Mutating nodes/neurons}
    \begin{itemize}
        \item A node/neuron can be added with a probability of 0.2, allowing for increased complexity of the resulting solution due to the increase in total parameters that our model needs to learn to converge. While this can prove helpful in some cases, in others it may cause overfitting of the dataset. 
        \item An existing node/neuron can be removed from within the architecture with a probability of 0.2, leading to simplification of the model by creating an optimal architecture with fewer neurons, consequently leading to lesser computational requirements. This is akin to dropout, implemented in conventional neural network training, allowing for better generalization.
    \end{itemize}
    \item \textit{Mutating edge connections}
    \begin{itemize}
        \item Creating a connection between two arbitrary nodes/neurons allows for increased precision that results in a well-fitted model due to faster convergence. The assigned probability for this was 0.5.
        \item The deletion of an edge connection, similar to deletion of a node, allows for increased generalizability of the model. In conventional machine learning, weight pruning serves a similar purpose, allowing for removal of less impactful weights in order to increase model efficiency with regards to size and performance. We set 0.5 as the probability for this happening.
    \end{itemize}
\end{itemize}

\subsubsection{Speciation}
In terms of optimizing for model topologies while training, newly mutated models could be at a disadvantage due to their non-convergent initial state. To give them a fair chance, NEAT implements speciation, which is the idea of dividing up the population based on topological and structural similarities. This division is decided based on the genomic distance between each genome which is calculated using the combination of how many non-homologous nodes and connections exist along with the divergence of homologous nodes and connections since their initialization. 

Upon division, the models work to improve their fitness by competing within their species, rather than against the whole population, allowing for nascent structures to be formed and optimized without being eliminated prematurely. This step is an analog to the parent and survivor selection step that is prevalent in many genetic algorithms. 

\section{Experiment \& Results}\label{Result}

\subsection{Overview}
Our experiment with regards to implementing NEAT focused on creating a 2D top-down car racing simulator, as shown in Fig. \ref{fig:3} and train it in such a way that it allowed for the agent car to loop around the track with reasonable driving stability while also not colliding with the boundary of the track itself. To ensure increases in model complexity, we created different map variations, each getting progressively more complex in terms of various parameter configurations.

\subsection{Simulator}
\begin{figure}[ht]
    \centering
    \includegraphics[width=0.45\textwidth]{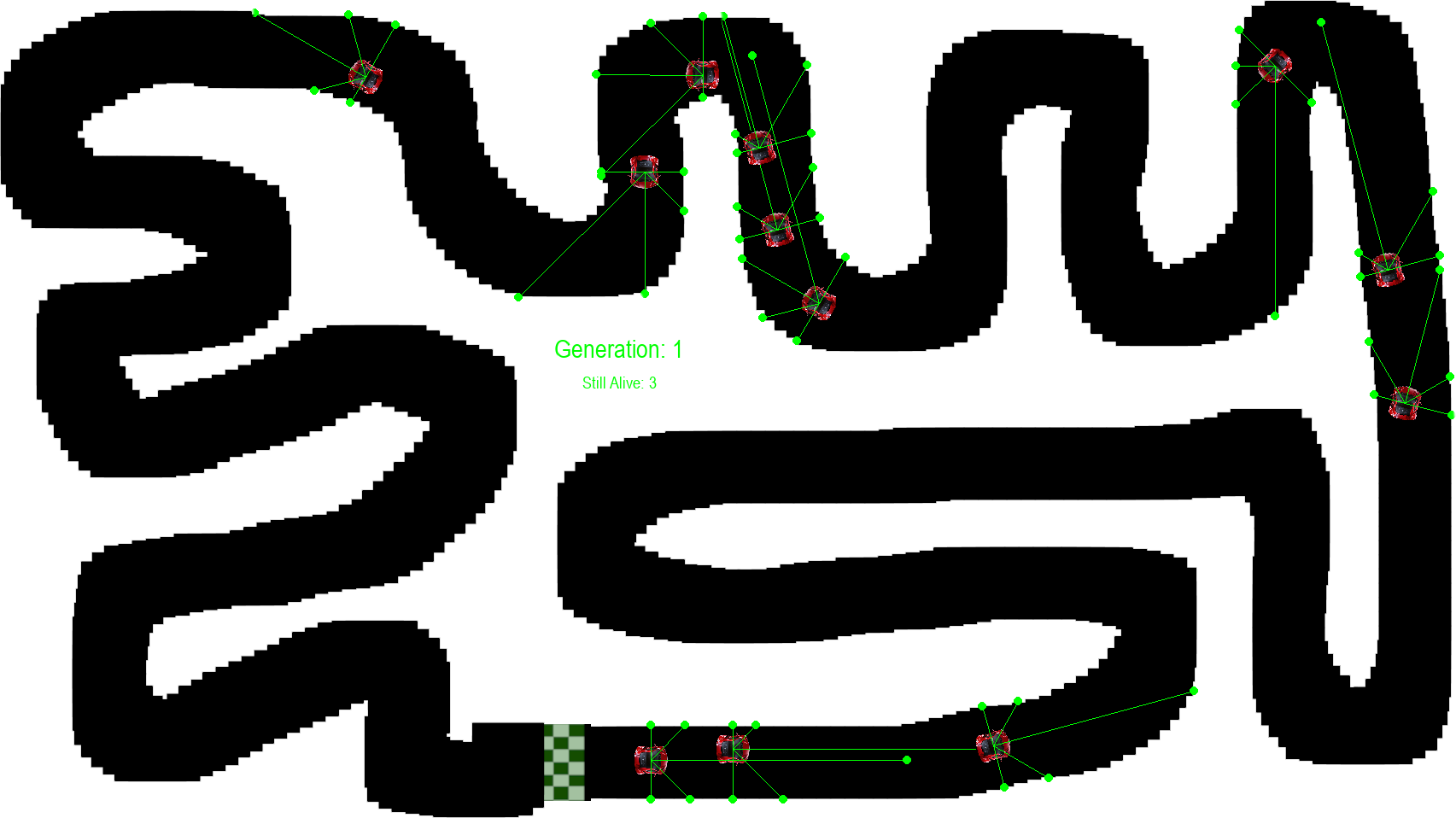}
    \caption{2D simulator built using Pygame, using map no. 4}
    \label{fig:3}
\end{figure}

Our simulator is a relatively simple top-down 2D track visualization, as shown in Fig. \ref{fig:3} in which our car sprites are superimposed on them, with their movement changing as our evolving neural network structures choose an action from a collection of possible actions that we will discuss when detailing the implementation of our car agent. 

We have implemented the ability to switch our maps to allow for a level of dynamism in the environment in which our neural networks are evolved to learn to loop the tracks. Our experiment focused on having six variations of the track that have changes incorporated in them based on the following aspects:

\begin{itemize}
    \item Total Distance
    \item Straight Sections
    \item Turn Sections
    \item Turn Sharpness
    \item Track Width
\end{itemize}

\begin{figure}[ht]
     \centering
     \subfloat[Simple loop]{%
        \label{fig:map1}
         \includegraphics[width=0.45\linewidth]{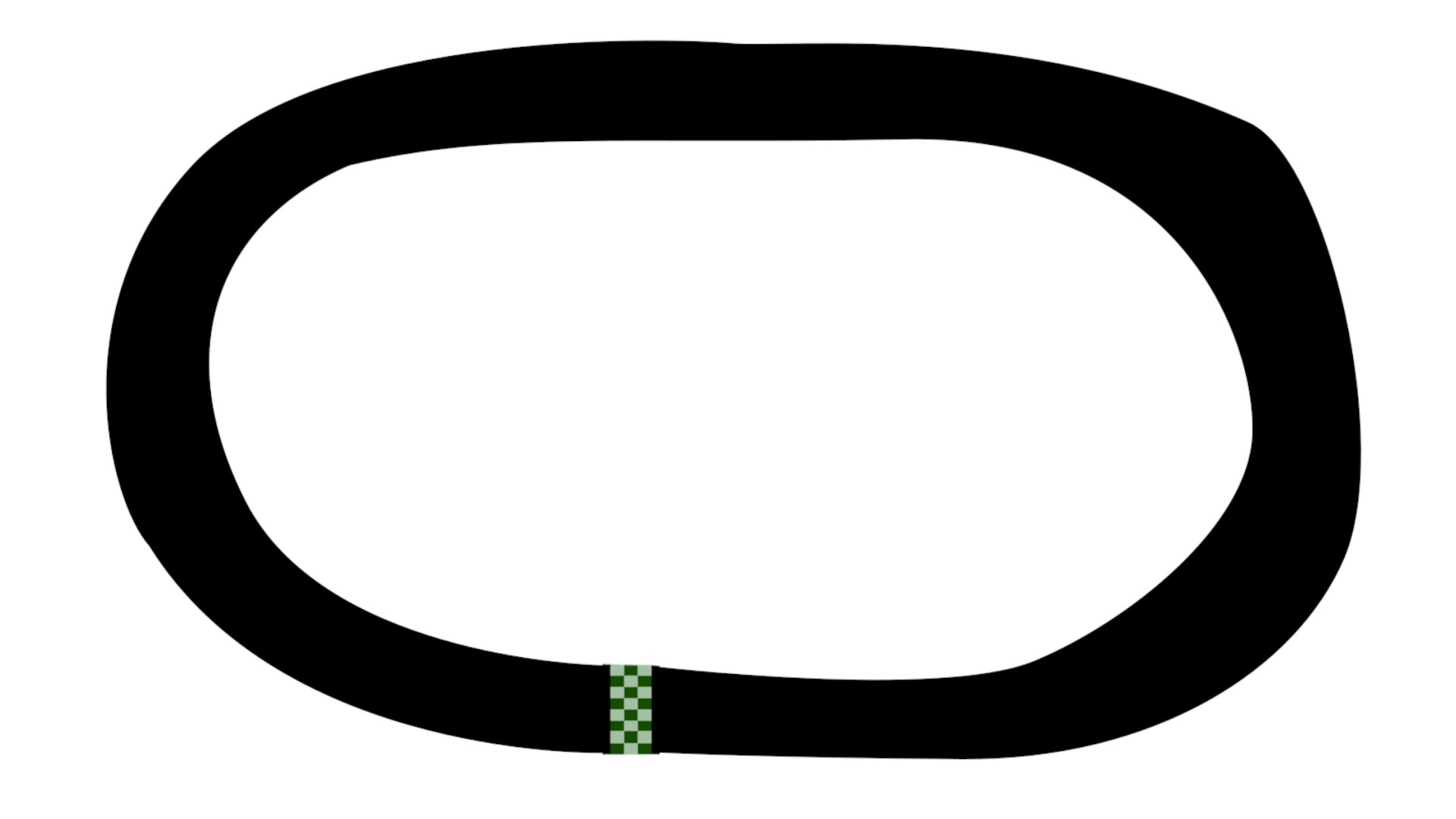}} 
     \hfill
     \subfloat[Loop with curves]{%
     \label{fig:map2}
         \includegraphics[width=0.45\linewidth]{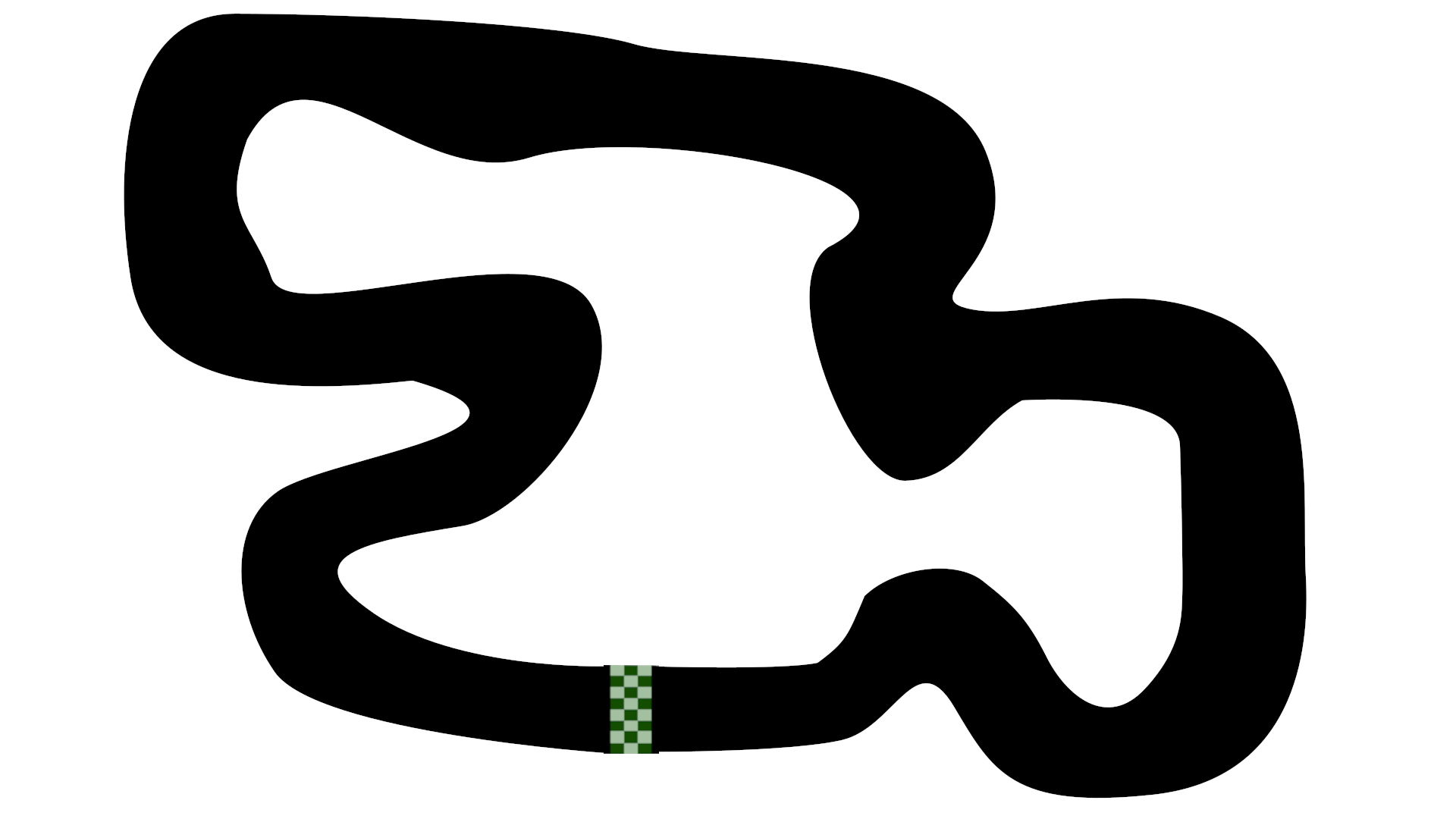}} 
     \subfloat[Slight sharp turns]{%
     \label{fig:map4}
         \includegraphics[width=0.45\linewidth]{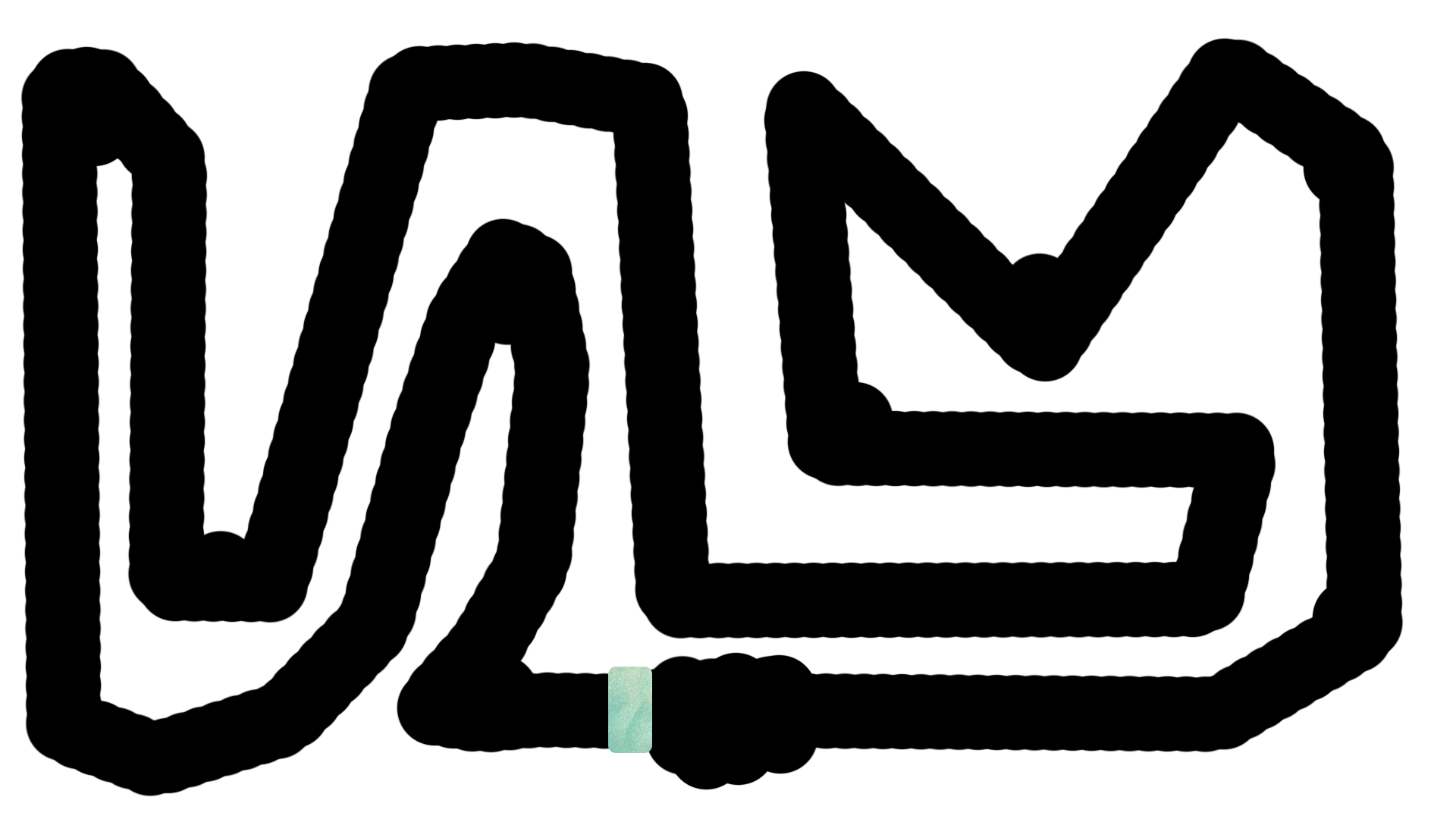}}
         \hfill
     \subfloat[Constant turns and twists]{%
     \label{fig:map3}
         \includegraphics[width=0.45\linewidth]{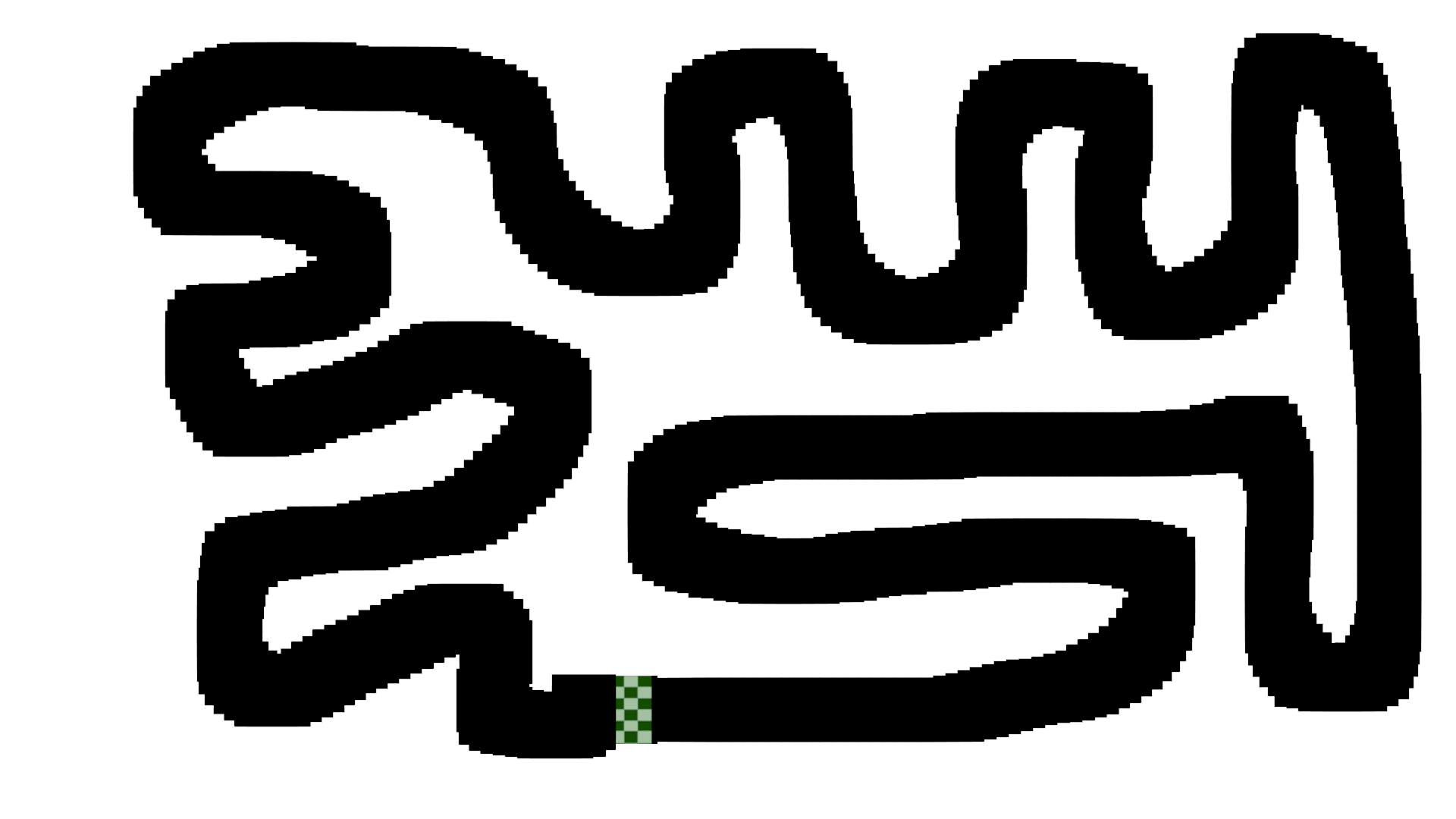}}
 
        \caption{Four different maps}
        \label{fig:4}
\end{figure}

Therefore, based on tweaks made to each of the aspects of the above mentioned aspects, we created a selection of six maps, depicted in Fig. \ref{fig:4}, that our NEAT algorithm will train on and generate neural network structures that perform well of all sorts of possible track configurations for increased generalization. 
\subsection{Car Configuration}
\begin{figure}[ht]
    \centering
    \includegraphics[width=0.3\textwidth]{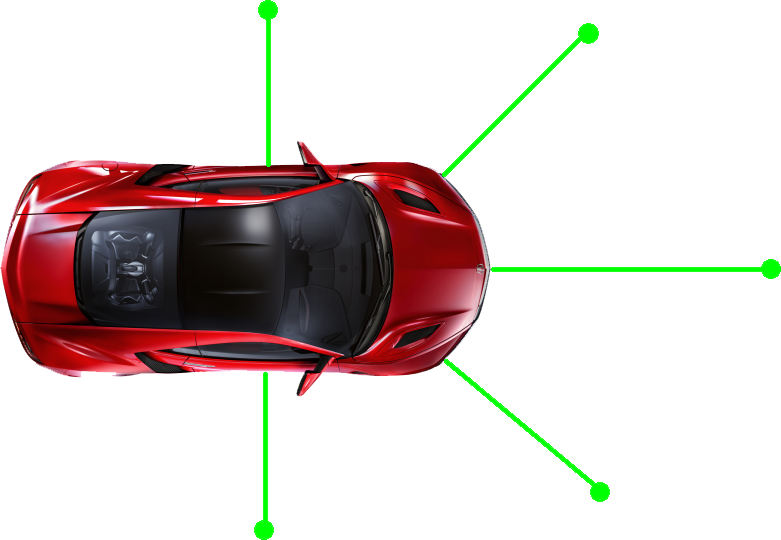}
    \caption{Car sprite with sensor visualization }
    \label{fig:5}
\end{figure}

For our experiment, the car serves as the agent that we use to represent the output and performance of the neural network that our NEAT algorithm is generating. Our car was represented by the sprite in Fig. \ref{fig:5}

At any given moment, our car agent is keeping track of the following:
\begin{itemize}
    \item \textit{Position}: This tracks the placement of the car with respect to its center. The representation of position is an $(x,y)$ coordinate which we calculate as follows:
        \begin{equation}
            angleRadian_{car} = rad(360-angle_{car})
        \end{equation}
        \begin{equation}
            xDist = \sin(angleRadian_{car})\times speed_{car} 
        \end{equation}
        \begin{equation}
            yDist = \cos(angleRadian_{car})\times speed_{car}
        \end{equation}
        \begin{equation}
            x= position_{car} + xDist
        \end{equation}
        \begin{equation}
            y= position_{car} + yDist
        \end{equation}
    The calculated $x$ and $y$ values are then clamped to ensure that they are not out-of-bounds.  
    \item \textit{Angle}: This value only increments and decrements in steps of 15 degrees only to ensure simplicity of control. The angle helps dictate the direction of the car on the track and allows it to turn accordingly.
    \item \textit{Speed}: This is to keep a record of how fast the car is moving across the track. The lower limit is clamped at 10 and the variation step size is $\pm$2.
    \item \textit{Distance}: This is a simple calculation of the area the car has covered over time. 
    \item \textit{Alive}: This is a boolean value that is returned as a result from our collision detection function. 
    \item \textit{Center of the agent}: This is calculated in order to cater to movement changes of our agent and serves as the origin point of our raycasts for our car `sensors'. 
    \item \textit{Radar}: This is a 5-array of `sensor' data which comprises of distance from the center of the car to the track boundary. We calculate this data for 5 different angles relative to the front of our car sprite, shown in Fig. \ref{fig:5}. The resulting 5-array is used as input for our NEAT algorithm which in turn is fed into the input layer of the various feed forward neural network structures that NEAT produces while training. 
\end{itemize}

\subsection{Q-Learning Parameters}

\begin{table}[ht]
    \caption{Main parameters of our Q-Learning implementation}
    \label{tab:Qparam}
    \centering
    \begin{tabular}{|c|c|}
         \hline
         \textbf{Parameter} & \textbf{Value} \\
         \hline
         Episodes (training) & 30,000\\
         Episodes (testing) & 10,000\\
         maxSteps & 2000\\
         epsilon & 0.8\\
         epsilonMin & 0.001\\
         learning rate & 0.8\\
         lrMin & 0.4\\
         gamma & 0.99\\
         \hline 
    \end{tabular}
\end{table}

Table \ref{tab:Qparam} shows the major parameters of our Q-Learning algorithm and the values that we chose for it, for our implementation. The reasoning behind all those value choices are as follows: 

\begin{itemize}
    \item \textit{Episodes (training)}: Decided through trial and error. The car showed some convergence, the car reached the final checkpoint and continued circling the track more times than not, and the size of the memory was also appropriate with $30,000$ episodes.
    \item \textit{Episodes (testng)}: $100$ episodes was sufficient to make an observation of the trained agent.
    \item \textit{maxSteps}: is chosen as 2000 to ensure that the episode does not stagnate as the car learns to stay on track.
    \item \textit{epsilon}: It determines the randomness to the action chosen at each step and so is set high at the start.
    \item \textit{epsilonMin}: To ensure a small amount of randomness even after training for a long time. This is done to make sure to not get stuck in local minima.
    \item \textit{learning rate}: The learning rate, which allows the Q-value to update, is initially set high so that the agent can learn rapidly.
    \item \textit{lrMin}: $lrMin$ is needed to ensure Q-value keeps updating and does not stagnate.
    \item \textit{gamma}: The discounting factor accounts for distant rewards in the q-value calculation.
\end{itemize}

Other than the parameters in Table \ref{tab:Qparam}, the checkpoints for each map were set discretely according to the need of each map.

\subsection{Q-Learning Results}
\begin{figure}[ht]
    \centering
    \includegraphics[width=\linewidth]{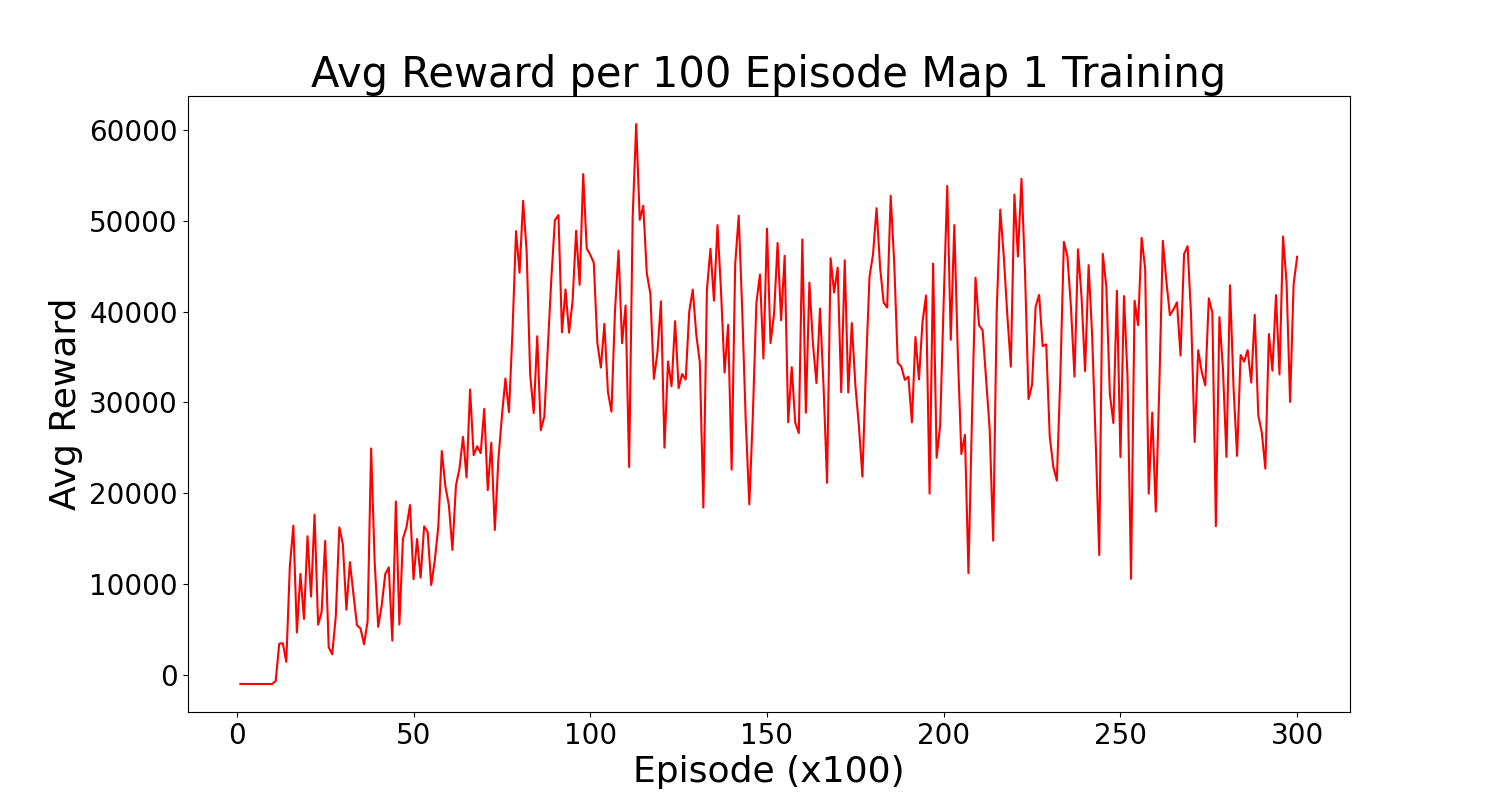}
    \caption{Average Reward per 100 Episodes during training Map 1}
    \label{fig:train1}
\end{figure}

\begin{figure}[ht]
    \centering
    \includegraphics[width=\linewidth]{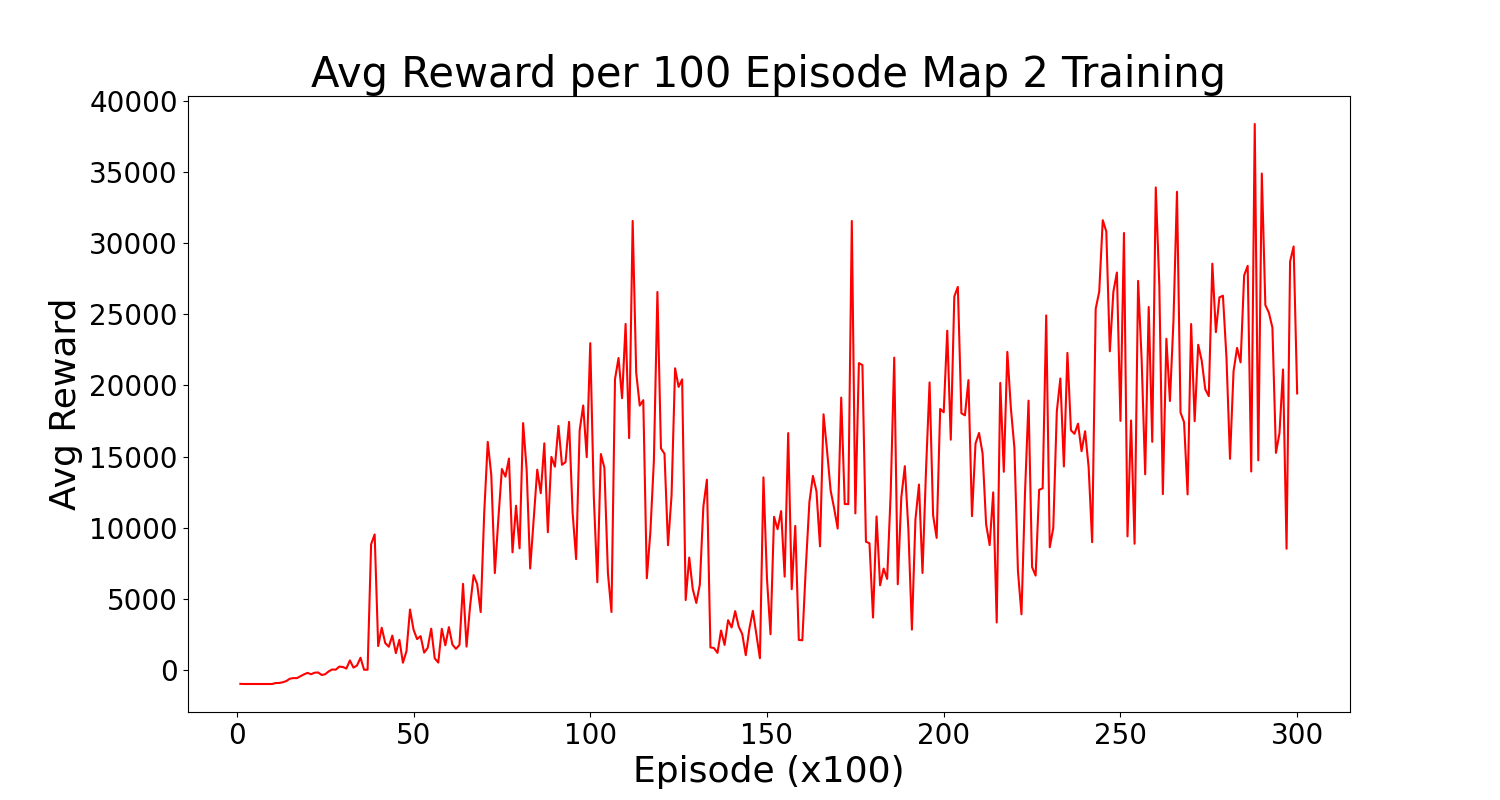}
    \caption{Average Reward per 100 Episodes during training Map 2}
    \label{fig:train2}
\end{figure}

\begin{figure}[ht]
    \centering
    \includegraphics[width=\linewidth]{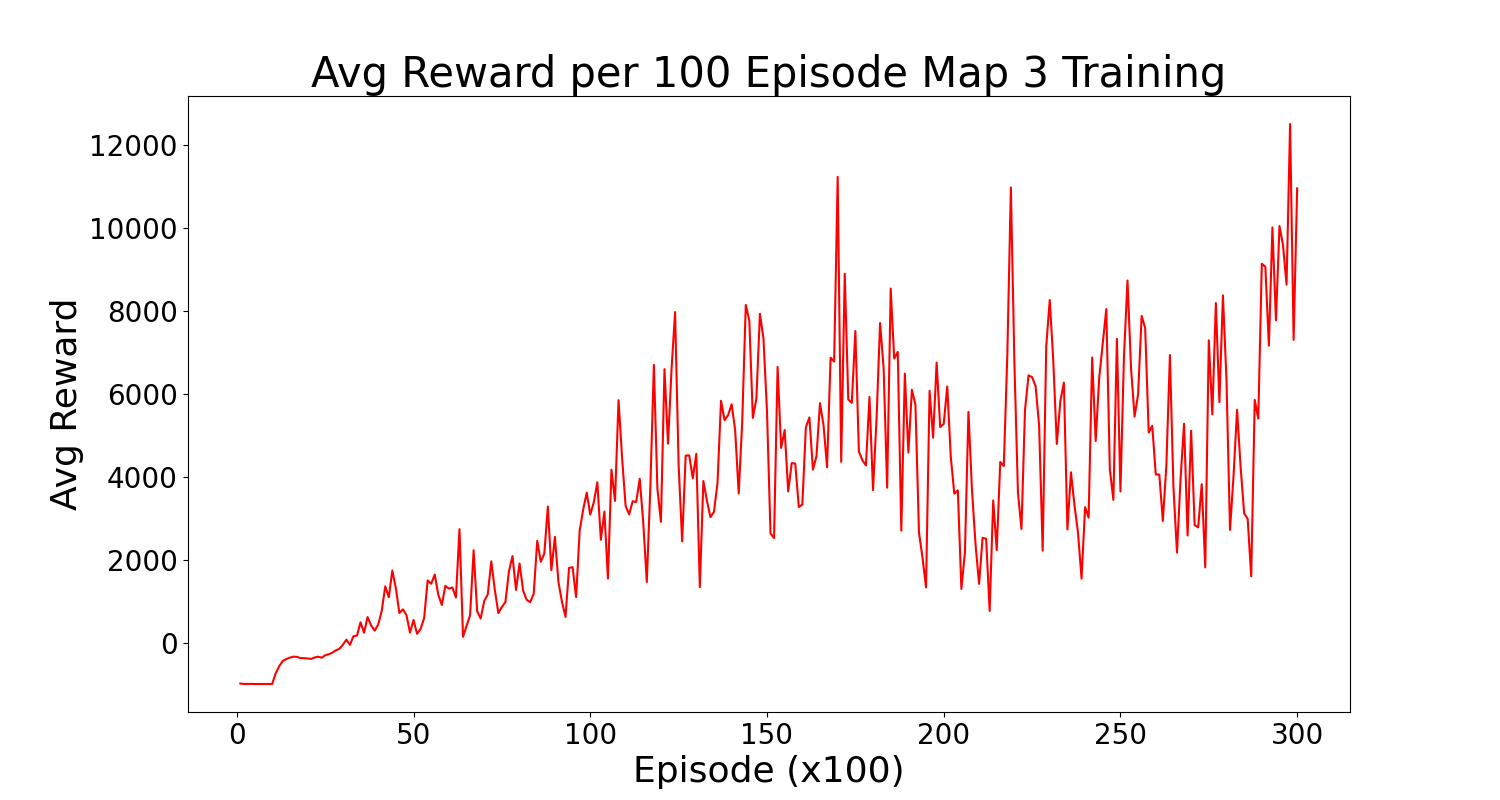}
    \caption{Average Reward per 100 Episodes during training  Map 3}
    \label{fig:train3}
\end{figure}

\begin{figure}[ht]
    \centering
    \includegraphics[width=\linewidth]{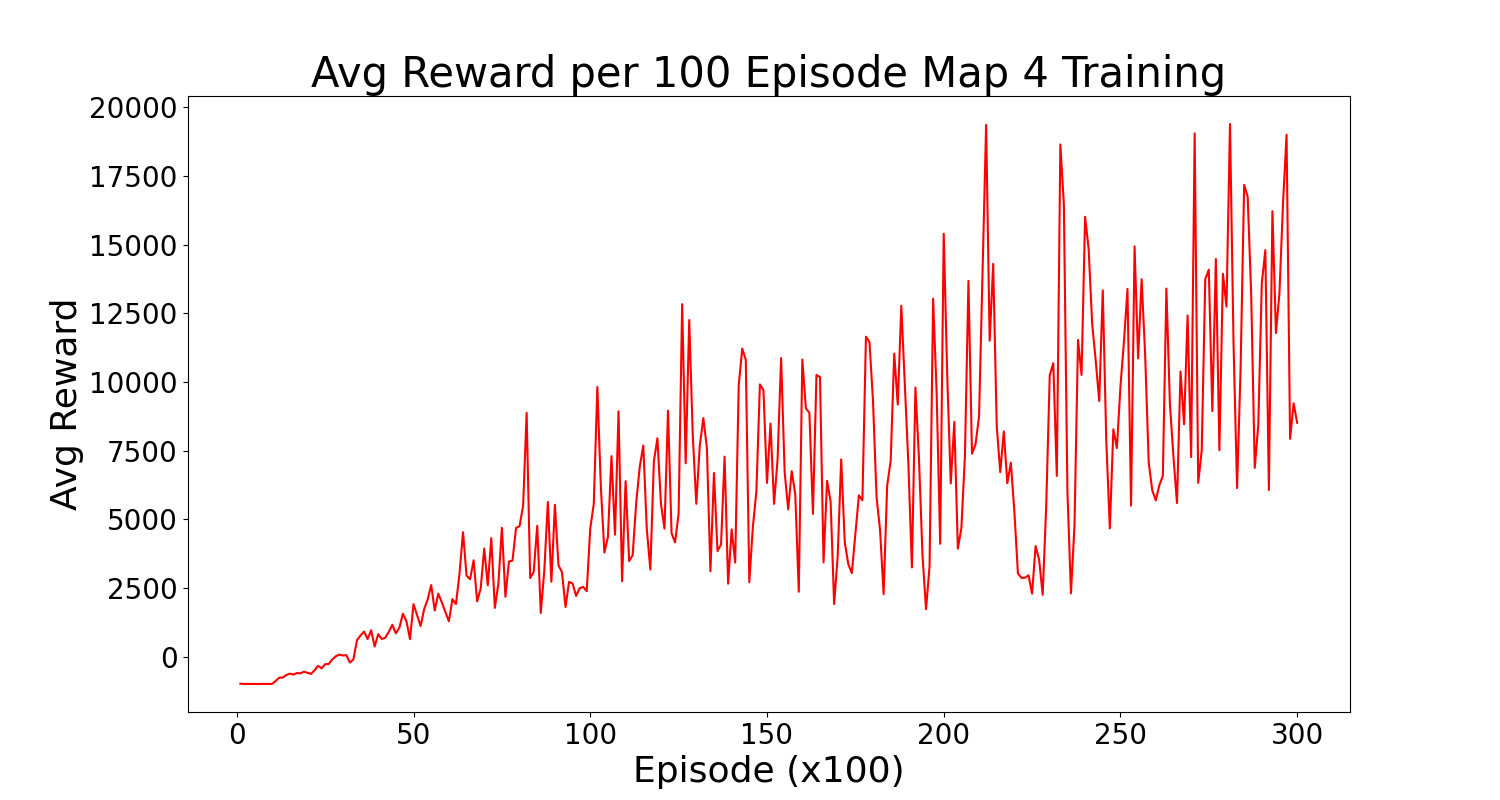}
    \caption{Average Reward per 100 Episodes during training  Map 4}
    \label{fig:train4}
\end{figure}

\begin{figure}[ht]
    \centering
    \includegraphics[width=\linewidth]{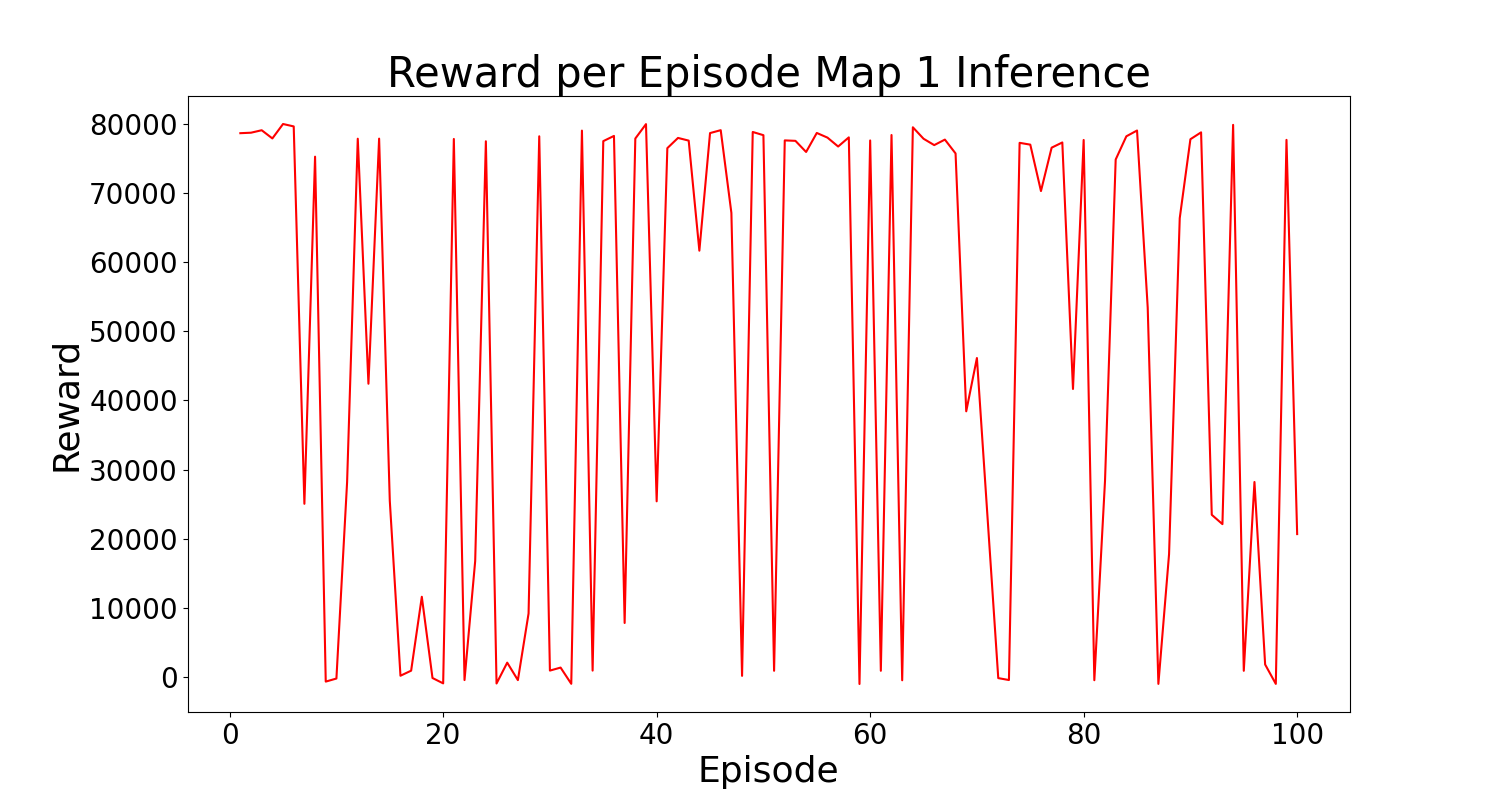}
    \caption{Reward per Episode during inference Map 1}
    \label{fig:test1}
\end{figure}

\begin{figure}[ht]
    \centering
    \includegraphics[width=\linewidth]{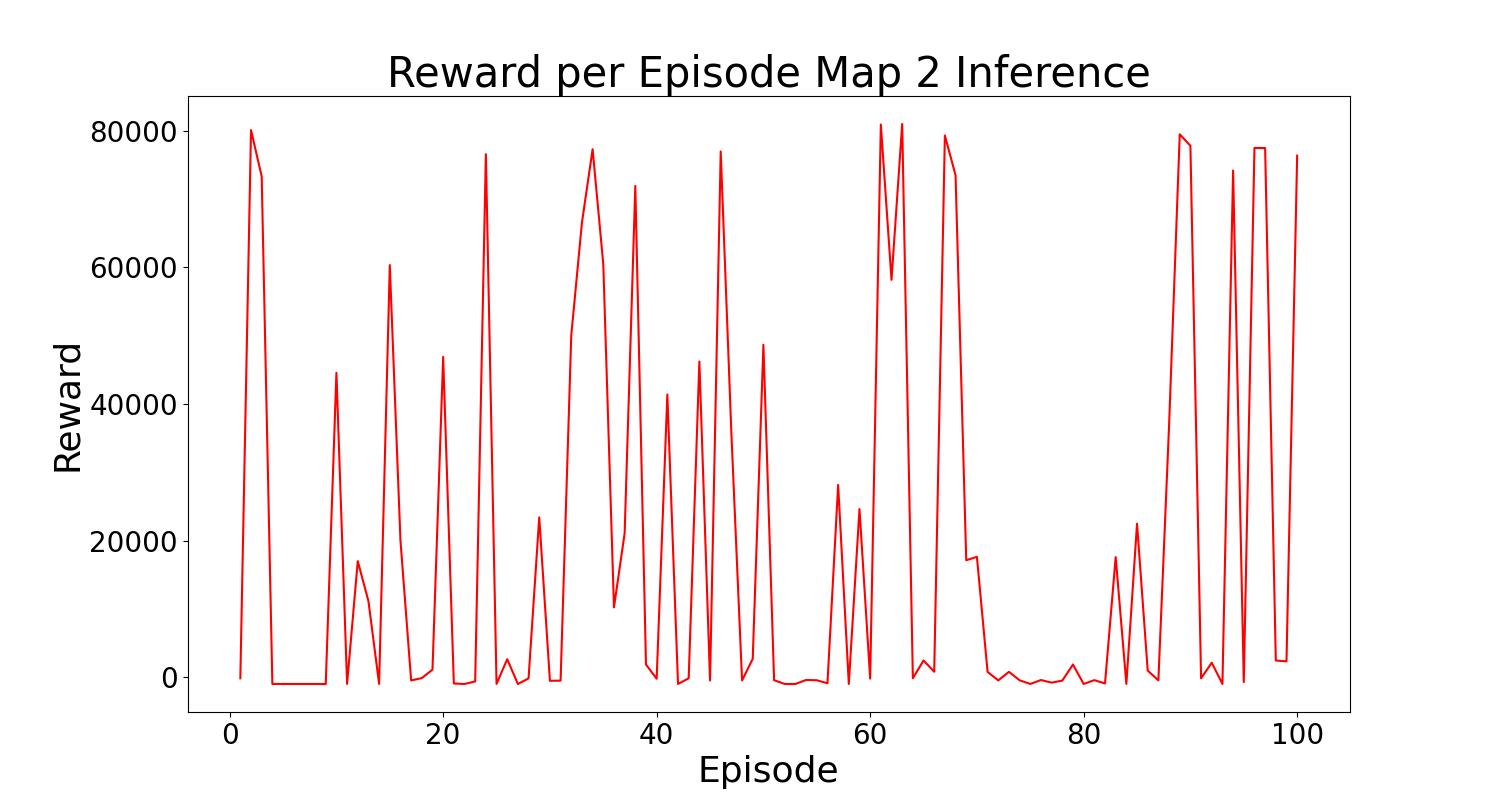}
    \caption{Reward per Episode during inference Map 2}
    \label{fig:test2}
\end{figure}

\begin{figure}[ht]
    \centering
    \includegraphics[width=\linewidth]{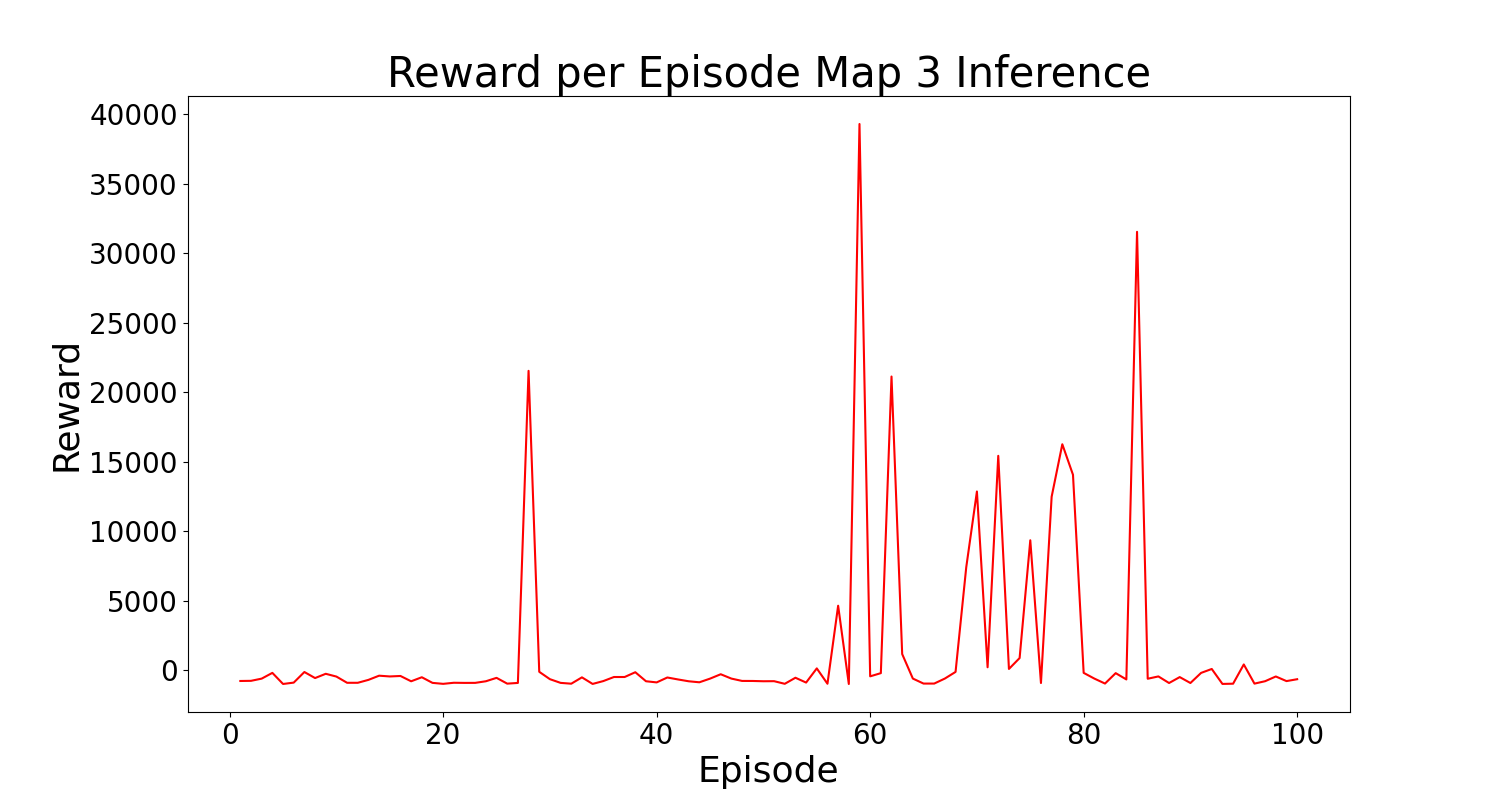}
    \caption{Reward per Episode during inference Map 3}
    \label{fig:test3}
\end{figure}

\begin{figure}[ht]
    \centering
    \includegraphics[width=\linewidth]{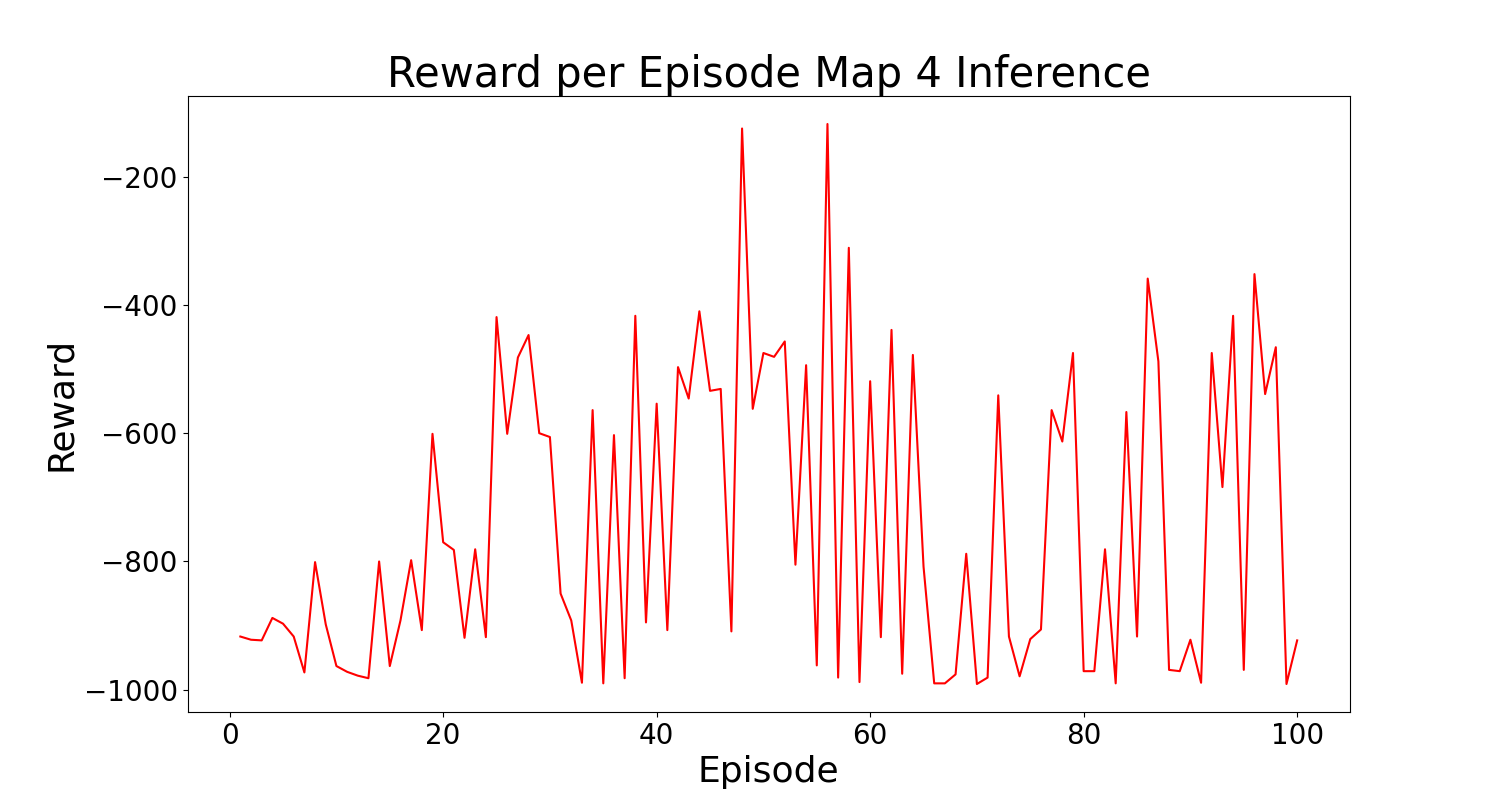}
    \caption{Reward per Episode during inference Map 4}
    \label{fig:test4}
\end{figure}

For each of the four maps, we ran the algorithm for 30,000 episodes each, taking the average reward ever 100 episodes we plot the average rewards. These reward plots can be seen in Fig. \ref{fig:train1} - \ref{fig:train4}. Furthermore, for each map we ran the algorithm on the trained car for 100 episodes to judge how well the car is performing post-training. The graphs for the rewards of the inference can be seen in Fig. \ref{fig:test1} - \ref{fig:test4}.

As we can see in Fig. \ref{fig:test1}, for the simple loop map (Fig.\ref{fig:4} \subref{fig:map1}), since there are not many turns and the path is rather simple, the car very quickly accumulates very high rewards. Within the first 10,000 episodes the average reward is as high as 50,000 and is steadily around 50,000 for the remaining episodes. Due to the simplicity of the map, despite having only four checkpoints, the car was able to get much higher rewards compared to the other maps which have more checkpoints. As for the trained car, we can see in Fig. \ref{fig:test1} that for mot only is the reward positive for most episodes, for a lot of them the reward is reaching as high as 80,000 points.

For the map with some curves (Fig. \ref{fig:4}\subref{fig:map2}), we can see that during the first 5000 episodes the average reward is growing steadily compared to the rapid growth in the simple loop. Moreover, due to the relative complexity, we can see that the maximum reward achieved id not as high. From \ref{fig:test2} we can see that although the trained car does achieve rewards up to 80,000, this reward is not achieved as often as in the simple loop map, despite having more checkpoints and thus more chances to get rewards in a each lap.

Map 3 (Fig. \ref{fig:4}\subref{fig:map3}) has some sharper turns and map 4 (Fig.\ref{fig:4} \subref{fig:map4}) has frequent turn and twists. We can see from Fig. \ref{fig:train3} and \ref{fig:train4} that the trajectory of rewards is similar, slower at the beginning and a steady increase. However we can see that during training for map 3 the reward is lower than for map 4. This may be due to the sudden and sharp turns compared to the smoother curves of map 4. However, during testing we can see that the car trained on map 3, Fig. \ref{fig:test3}, has better rewards than the one trained on map 4, Fig. \ref{fig:test4}. In fact we can see that all rewards from map 4 are negative which means the car was unable to even get past a few curves whereas for map 4 although few we do see some positive rewards.

\subsection{NEAT Parameters}

\begin{table}[ht]
\caption{Main parameters of our NEAT implementation}
    \label{tab:NEATparam}
    \centering
    \begin{tabular}{|c|c|}
         \hline
         \textbf{Parameter} & \textbf{Value} \\
         \hline
         Fitness Criteria & max\\
         Population Size & 200\\
         Total Generations & 100\\
         Activation Function & tanh\\
         Activation Mutation Rate & 0.02\\
         Node Mutation Rate & 0.2\\
         Connection Mutation Rate & 0.5\\
         Maximum Stagnation & 10\\
         Species Elitism & 1\\
         Species Fitness Criteria & max\\
         Population Elitism & 2\\
         \hline 
    \end{tabular}
\end{table}

Table \ref{tab:NEATparam} shows the major parameters of our NEAT algorithm and the values that we chose for it, for our implementation. The reasoning behind all those value choices are as follows: 

\begin{itemize}
    \item \textit{Fitness Criteria}: Just like most optimization problems, our goal was to maximize our fitness, thus we kept it that.
    \item \textit{Population Size}: To allow for a healthy diversity and increased number of species while also keeping in mind computational limits of our available hardware, we decided on a population size of 200 genomes as a reasonable amount.
    \item \textit{Total Generations}: Each run of NEAT trained for 100 generations to allow for enough time where exploration via mutation and exploitation via survivor selection would yield well converged structures. 
    \item \textit{Activation Function}: Our choice here was a bit arbitrary, in the sense that we could have either used the typical $sigmoid$ function or the chosen $tanh$ function since either of them allow to repsent the output between a certain value, which in the case of $tanh$ will be 1 to -1, allowing for more intuitive mapping of the output of this function to the action taken by our car agent. 
    \item \textit{Mutation Rates:}
    \begin{itemize}
        \item \textit{Activation}: We did not want for the $tanh$ activation function to deviate too far from its functionality, therefore we kept the mutation probability to 0.02.
        \item \textit{Node}: While we wanted to allow NEAT to explore new neural network structures, we also wanted to ensure that it does not deviate too far from our preassigned layer sizes, therefore we kept the mutation probability at 0.2.
        \item \textit{Connections}: To ensure quick convergence with less overfitting, we assigned equal probabilities to both the addition and removal of connections i.e. 0.5.
        \item \textit{Maximum Stagnation}: Stagnation is the idea of a genome's or specie's fitness not improving over generations. The maximum stagnation paramter allows for removal of such genomes and species that do not show any improvement for a predetermined number of generations. We chose 10 generations as the value for this.
        \item \textit{Species Elitism}: This is the same as the general concept of elitism as part of survival selection but is applicable within species as opposed to the whole population. We chose the value of 1 to allow for only the best genome of a specie to progress further into the next generation. 
        \item \textit{Species Fitness Criteria}: This allowed for us to decide how a species as a whole would be compared against other species. Using the highest fitness in a specie to represent the specie fitness is most ideal for cases where we need to converge fast and only want to optimize for the best possible genome whereas, we can use a mean of all genome fitness values when we need the species as a whole to improve in performance. We went for max since we required the best possible genome for our car simulation. 
        \item \textit{Population Elitism}: We chose 2 as we just needed a single genome to serve as a starting point for our next generation.  
    \end{itemize}
\end{itemize}

\begin{figure}[ht]
    \centering
    \includegraphics[width=\linewidth]{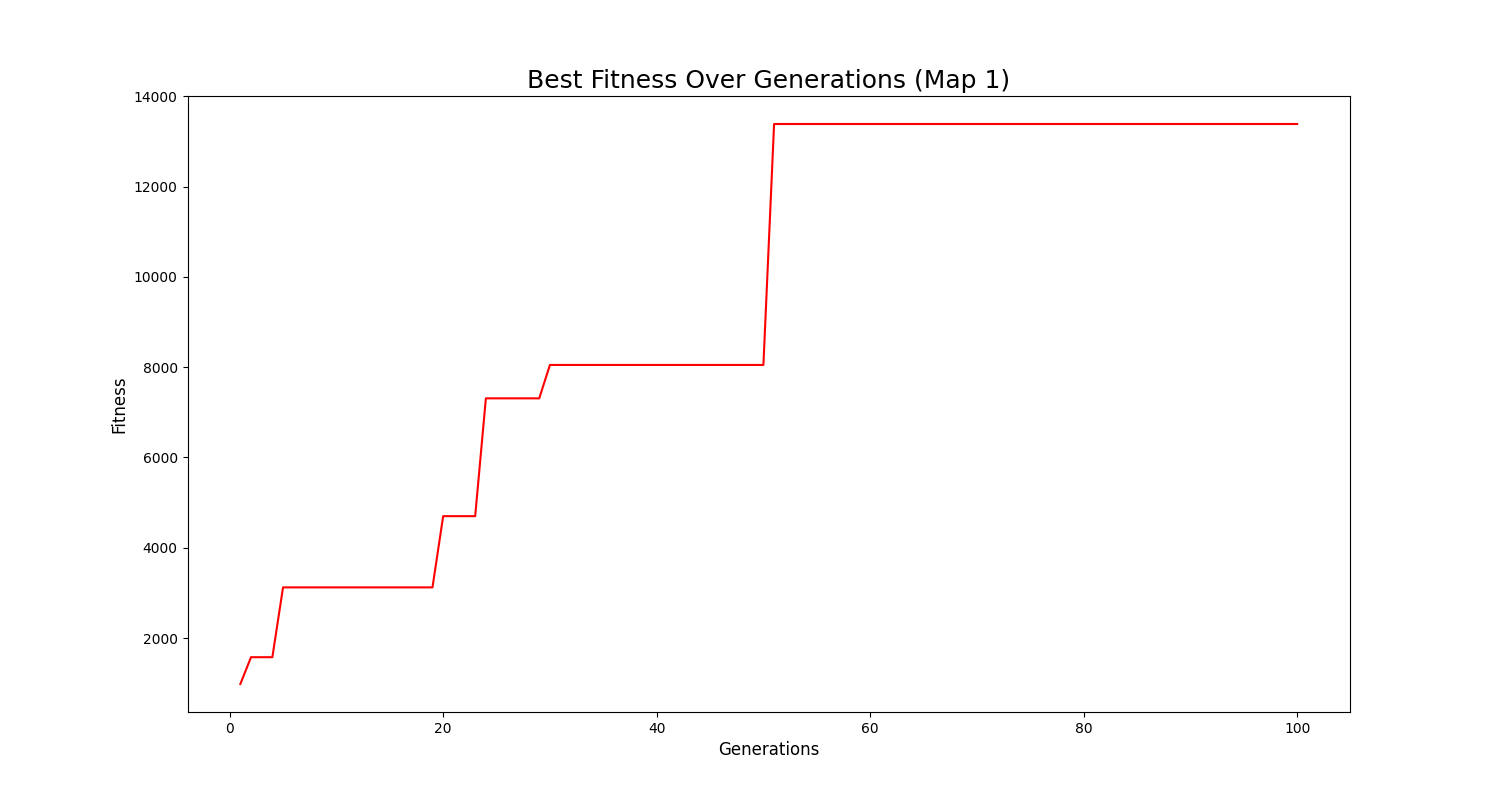}
    \caption{Best Fitness Over Generations Map 1}
    \label{fig:bfg1}
\end{figure}

\begin{figure}[ht]
    \centering
    \includegraphics[width=\linewidth]{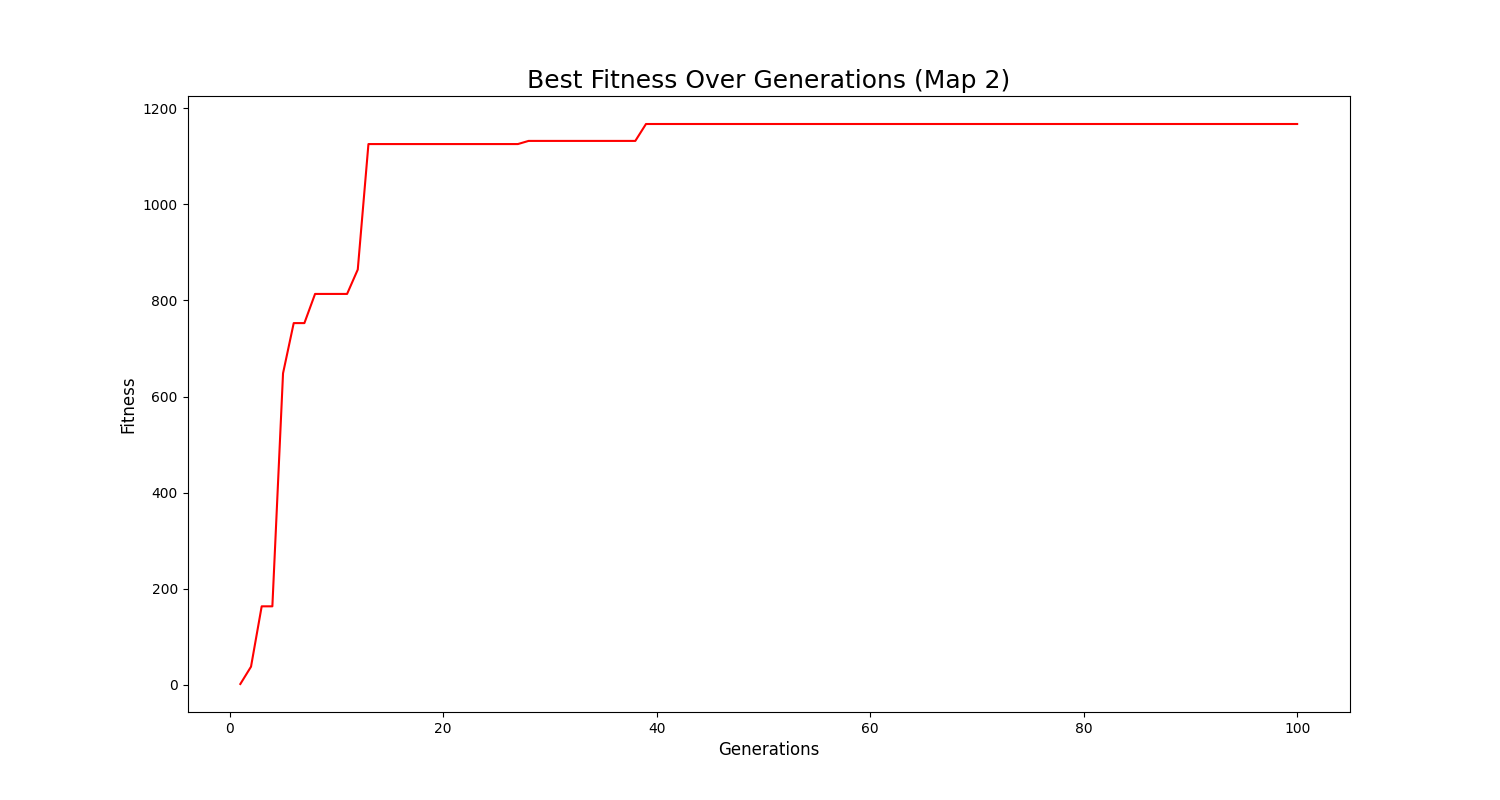}
    \caption{Best Fitness Over Generations Map 2}
    \label{fig:bfg2}
\end{figure}

\begin{figure}[ht]
    \centering
    \includegraphics[width=\linewidth]{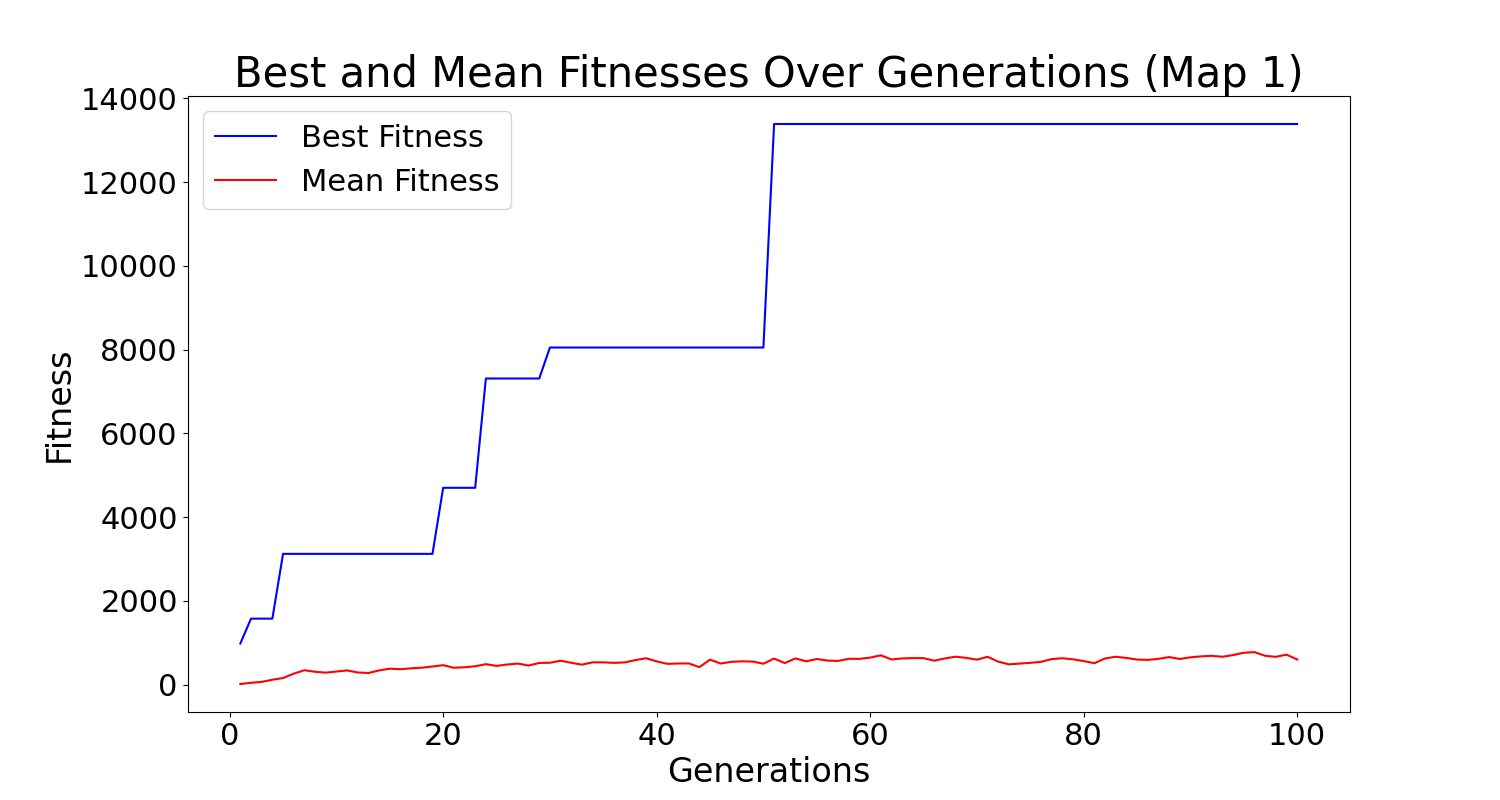}
    \caption{Best and Mean Fitness Over Generations Map 1}
    \label{fig:bmfg1}
\end{figure}

\begin{figure}[ht]
    \centering
    \includegraphics[width=\linewidth]{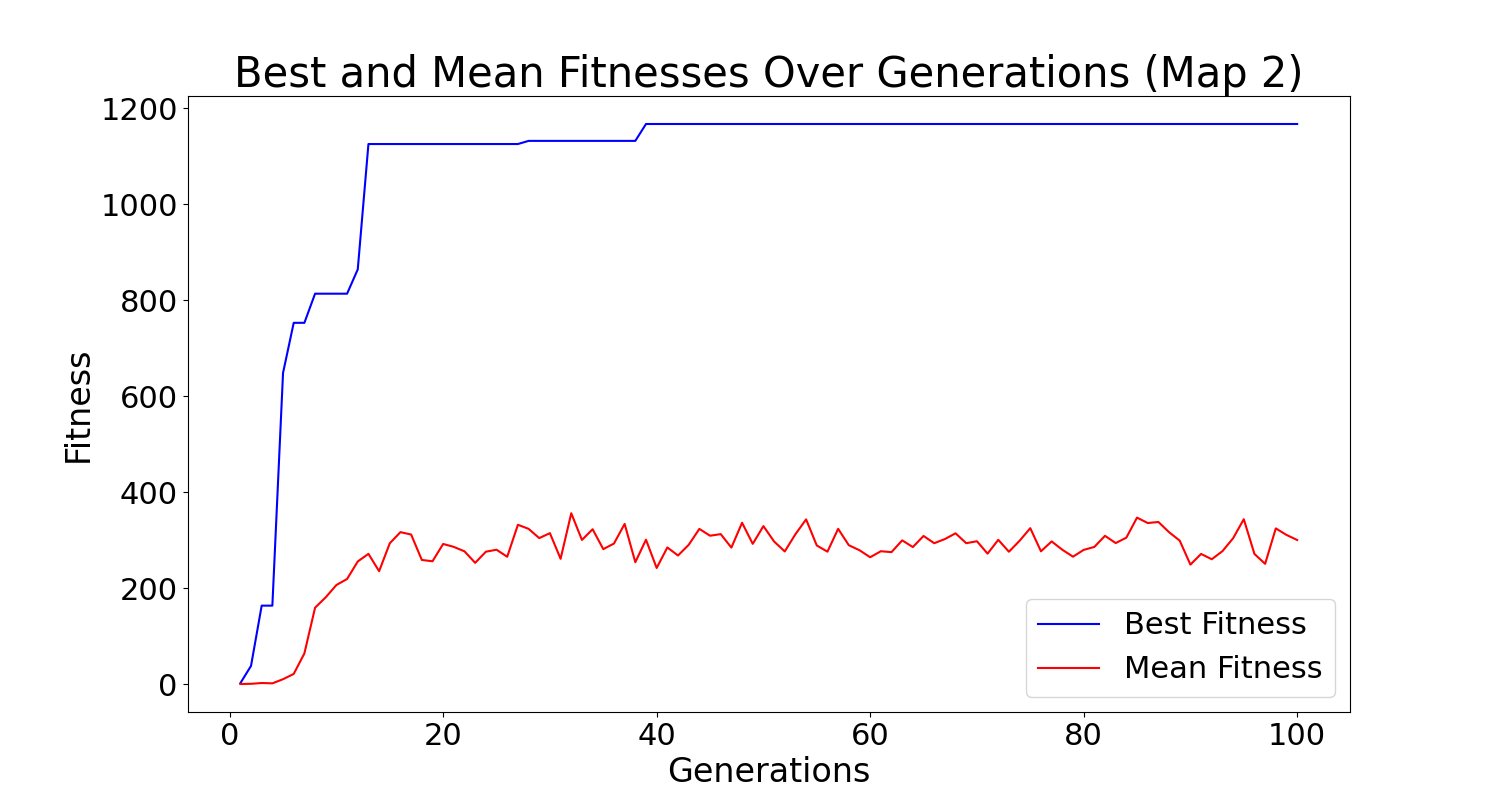}
    \caption{Best and Mean Fitness Over Generations Map 2}
    \label{fig:bmfg2}
\end{figure}

\begin{figure}[ht]
    \centering
    \includegraphics[width=\linewidth]{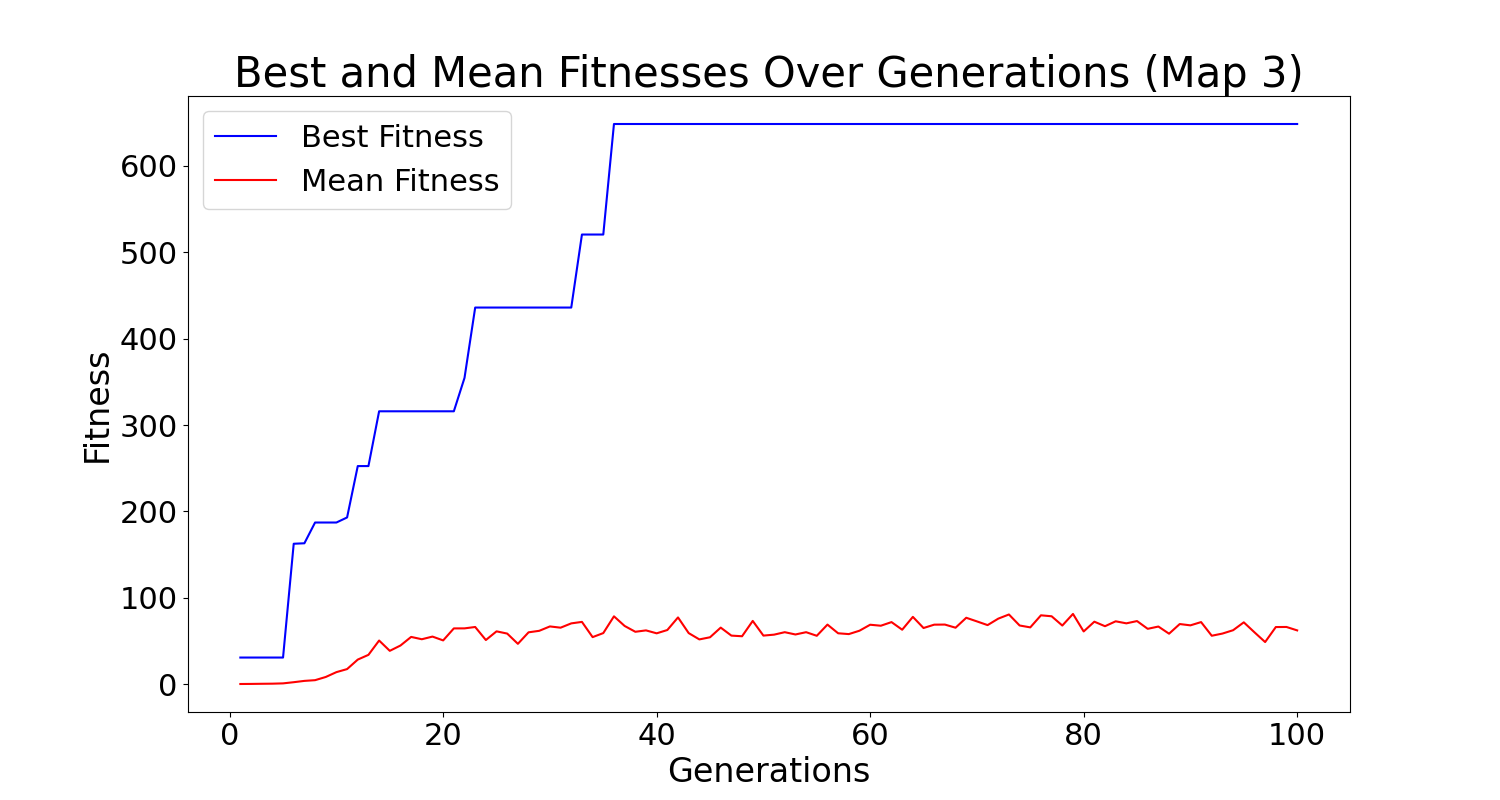}
    \caption{Best and Mean Fitness Over Generations Map 3}
    \label{fig:bmfg3}
\end{figure}

\begin{figure}[ht]
    \centering
    \includegraphics[width=\linewidth]{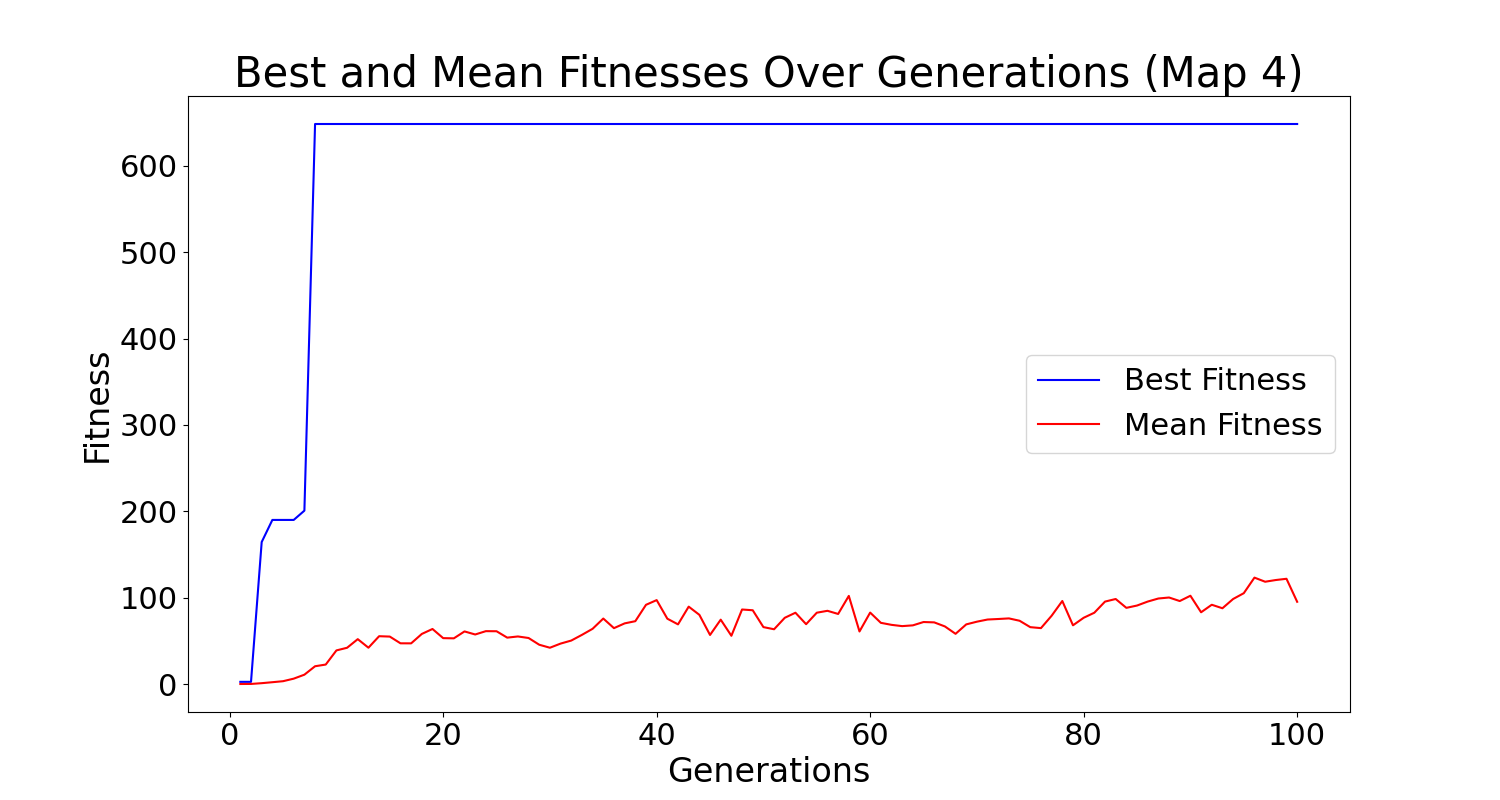}
    \caption{Best Fitness Over Generations Map 4}
    \label{fig:bmfg4}
\end{figure}

\begin{figure}[ht]
    \centering
    \includegraphics[width=\linewidth]{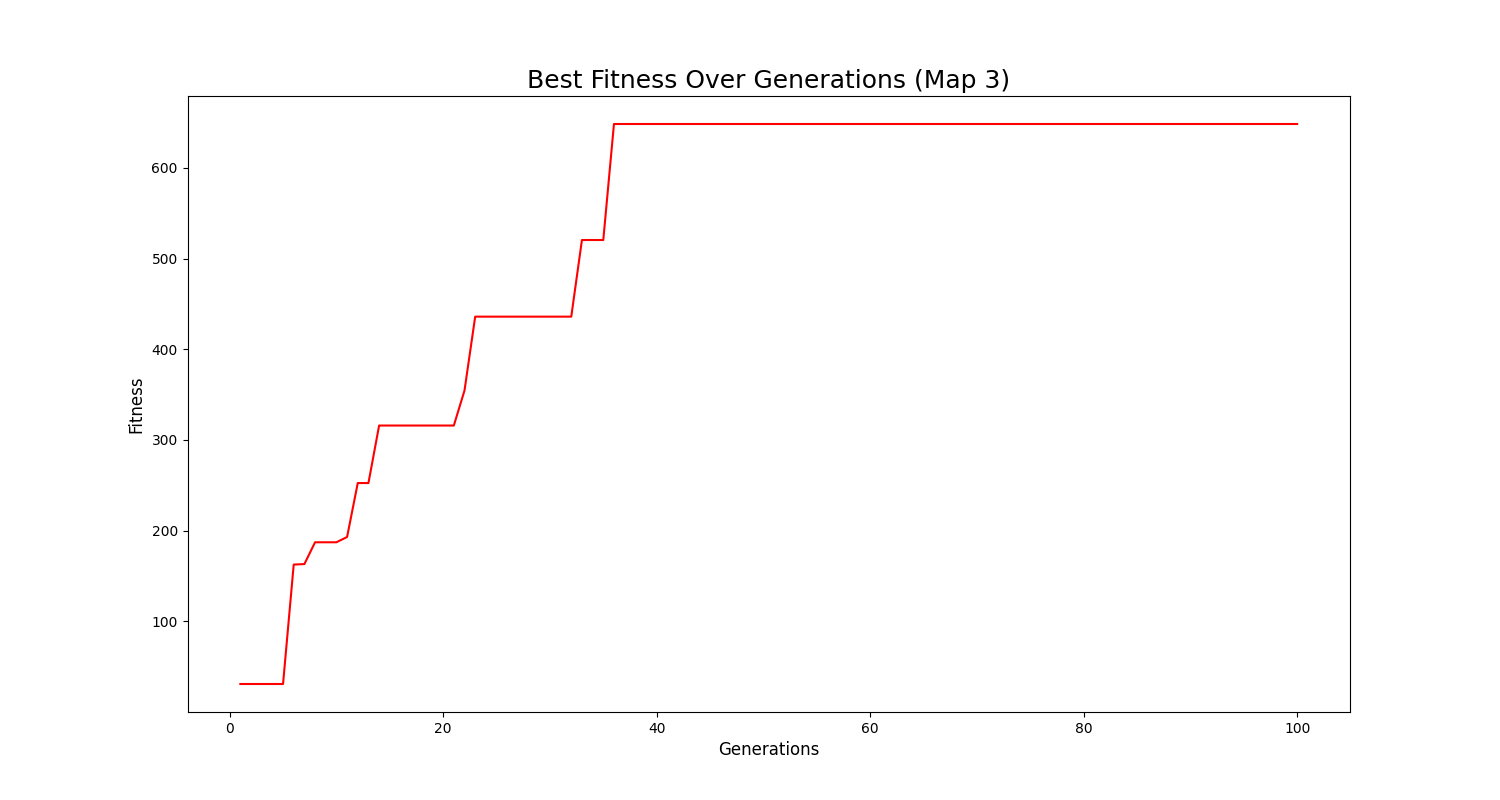}
    \caption{Best Fitness Over Generations Map 3}
    \label{fig:bfg3}
\end{figure}

\begin{figure}[ht]
    \centering
    \includegraphics[width=\linewidth]{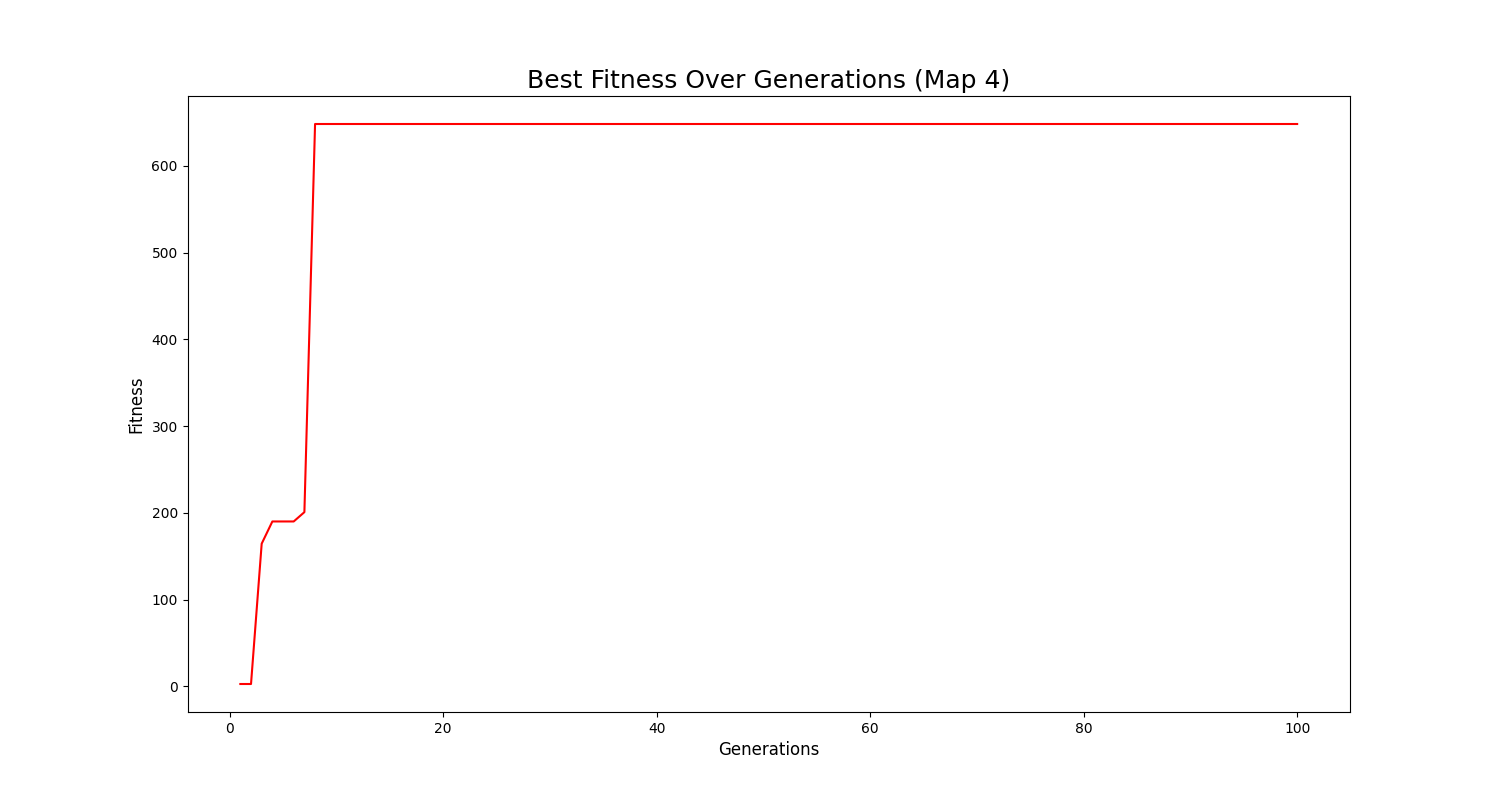}
    \caption{Best Fitness Over Generations Map 4}
    \label{fig:bfg4}
\end{figure}

After the initial setup of the experiment, we ran the algorithm for 100 generations on each of the four maps. We can see in Fig \ref{fig:bfg1}-\ref{fig:bfg4}, the best fitness that the model reached throughout 100 generations.

For the first map Fig.\ref{fig:4} \subref{fig:map1} which is a simple loop we can see in graph in Fig.\ref{fig:bfg1} that the model reaches it's optimal fitness which is about 140000 in just 50 generations. The straight line after 50 generations represent that the model wasn't able to further improve because of the limited scope of learning in a straightforward map with few turns. From Fig\ref{fig:bfg4} and Fig.\ref{fig:bfg3} we can see that the the model converged to the optimal results in just a few generations, for maps\ref{fig:4} \subref{fig:map3} and \ref{fig:4} \subref{fig:map4}, but the optimal fitness is around 600 which is significantly lower than that of the previous maps. This is due to the sharp turns and edges on map.\ref{fig:4} \subref{fig:map3} and \ref{fig:4} \subref{fig:map4}. 

\begin{figure}[ht]
    \centering
    \includegraphics[width=\linewidth]{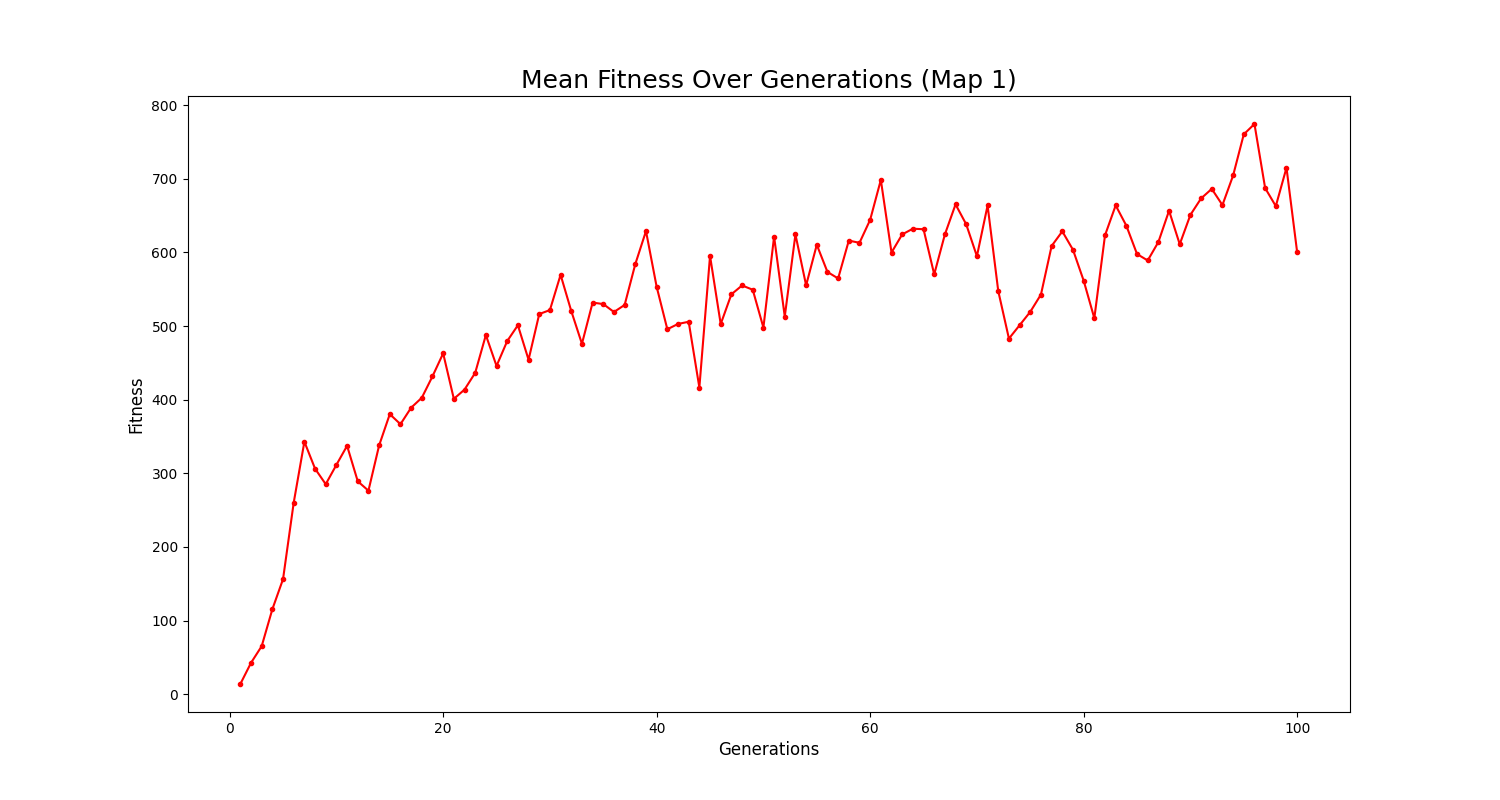}
    \caption{Mean Fitness Over Generations Map 1}
    \label{fig:mfg1}
\end{figure}

\begin{figure}[ht]
    \centering
    \includegraphics[width=\linewidth]{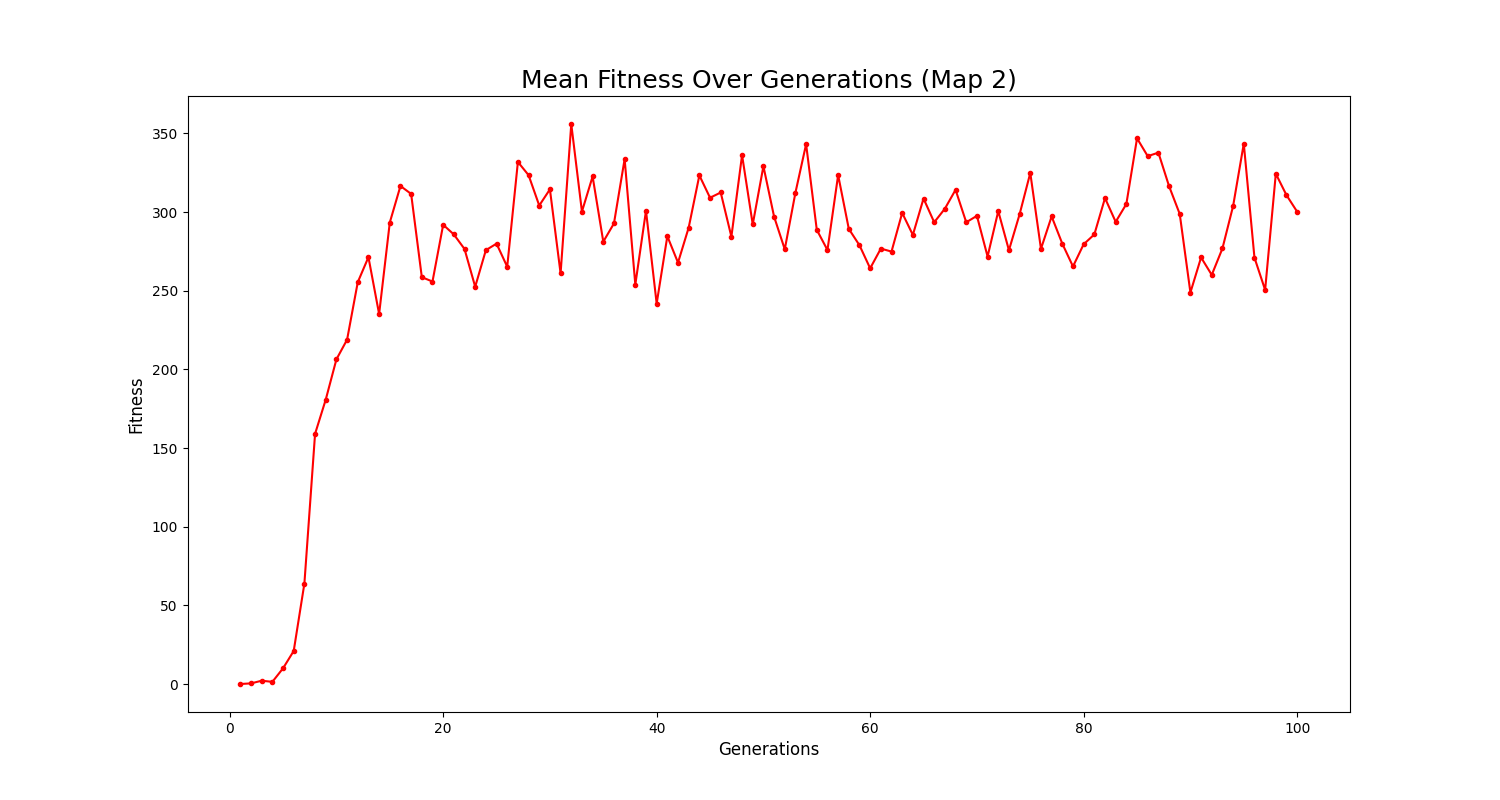}
    \caption{Mean Fitness Over Generations Map 2}
    \label{fig:mfg2}
\end{figure}

\begin{figure}[ht]
    \centering
    \includegraphics[width=\linewidth]{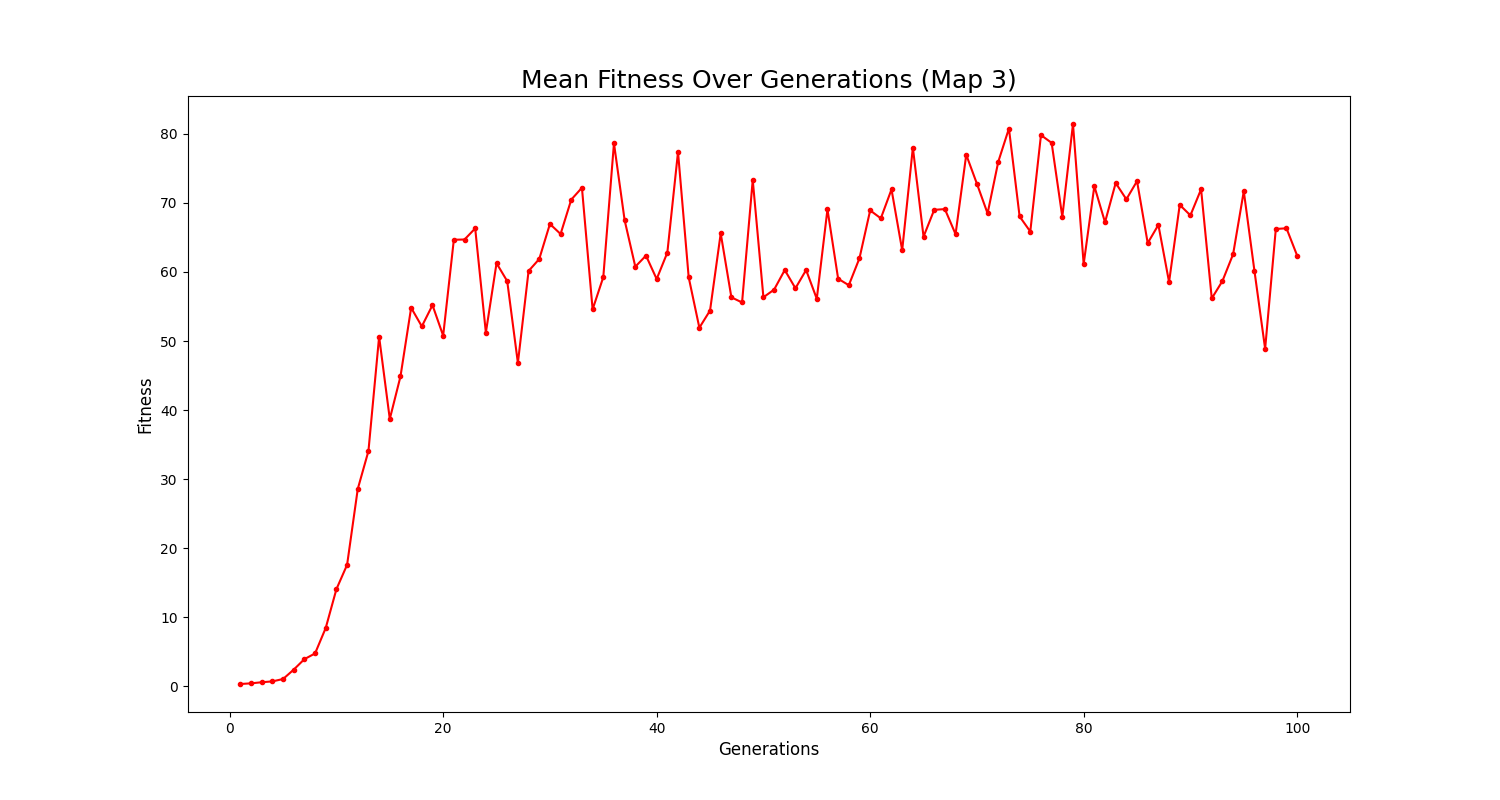}
    \caption{Mean Fitness Over Generations Map 3}
    \label{fig:mfg3}
\end{figure}

\begin{figure}[ht]
    \centering
    \includegraphics[width=\linewidth]{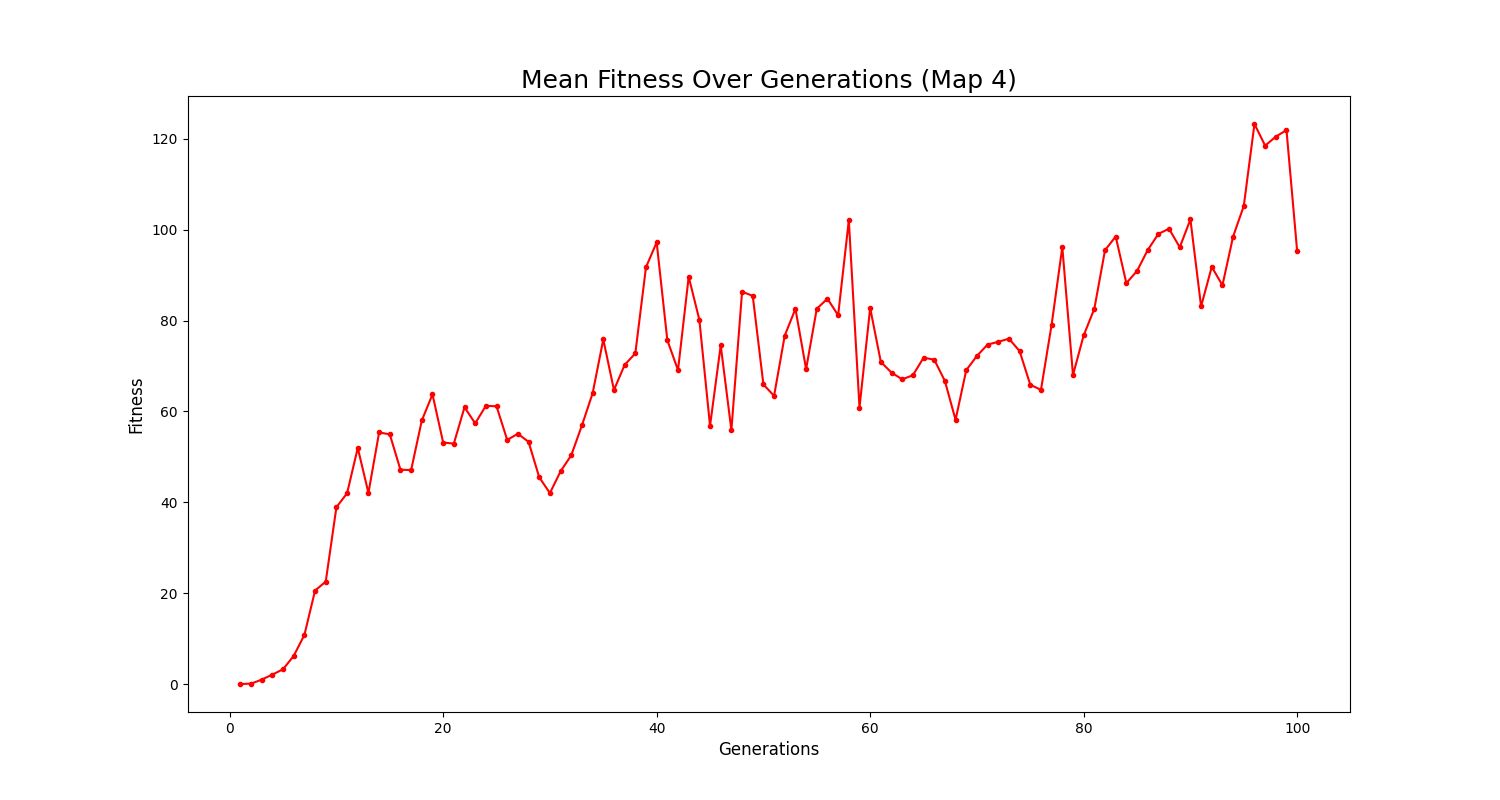}
    \caption{Mean Fitness Over Generations Map 4}
    \label{fig:mfg4}
\end{figure}

From Fig\ref{fig:mfg1}-\ref{fig:mfg4} represent the mean score of models on each of the four maps. For the first map as seen in fig\ref{fig:mfg1} the growth in the mean fitness over the generations is steady and reaches it's highest peak at around 90 generations. The trend in fig.\ref{fig:mfg2} and fig.\ref{fig:mfg3} increases slowly at the start but reach it's highest peak near 30-35 generations. For map.\ref{fig:4} \subref{fig:map4} 

\begin{figure}[ht]
    \centering
    \includegraphics[width=\linewidth]{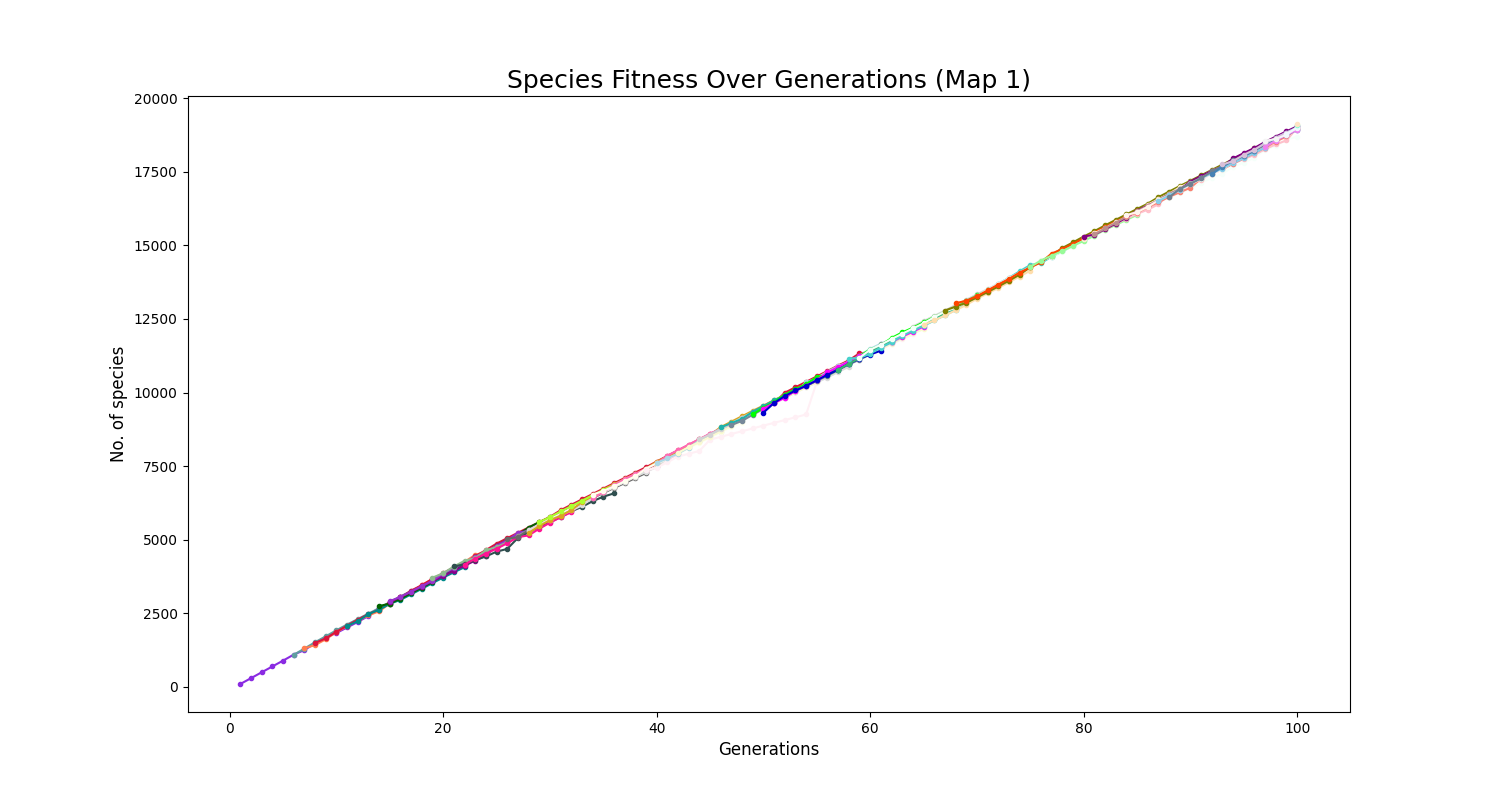}
    \caption{Species Fitness Over Generations Map 1}
    \label{fig:sfg1}
\end{figure}

\begin{figure}[ht]
    \centering
    \includegraphics[width=\linewidth]{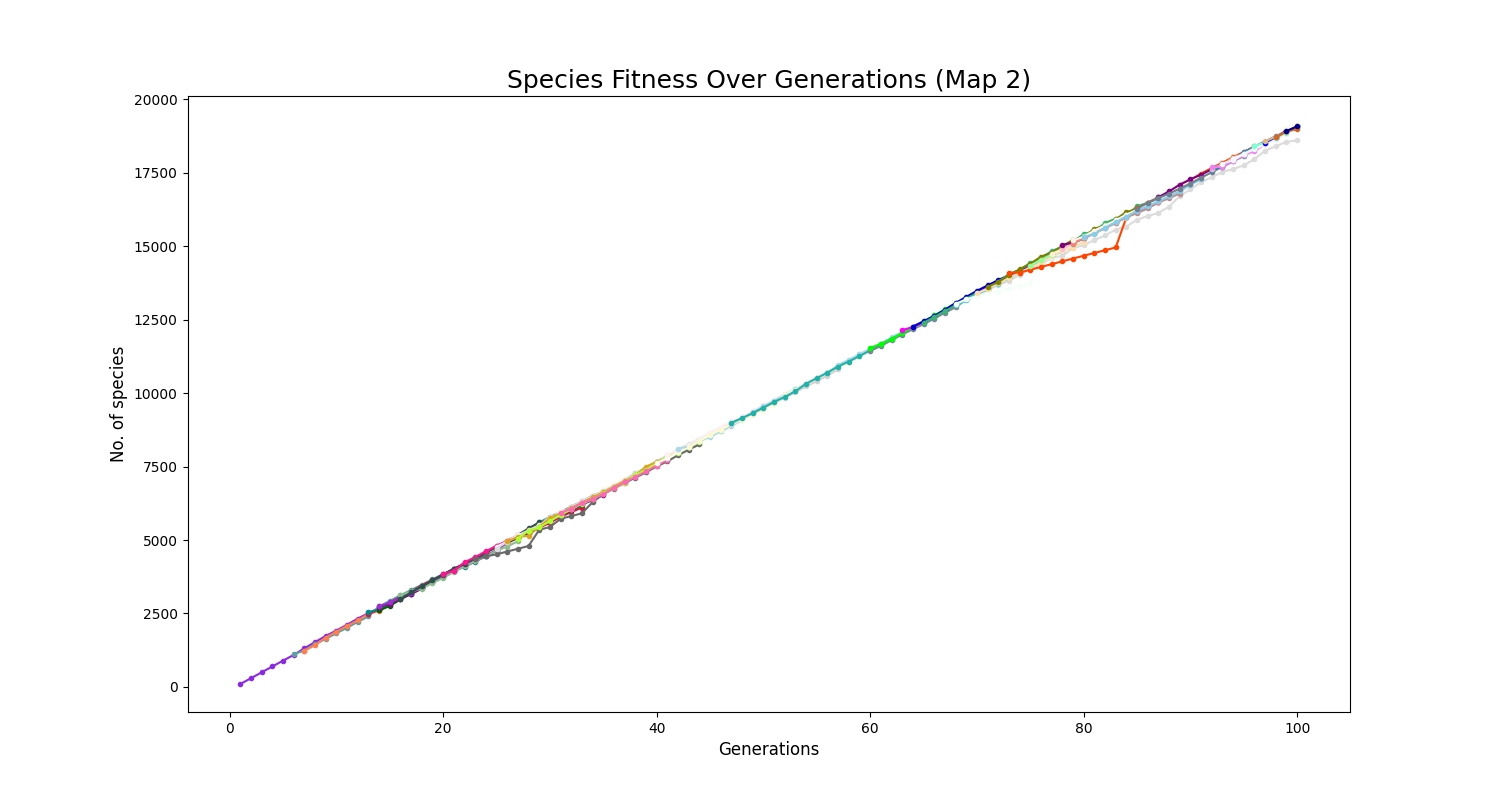}
    \caption{Species Fitness Over Generations Map 2}
    \label{fig:sfg2}
\end{figure}

\begin{figure}[ht]
    \centering
    \includegraphics[width=\linewidth]{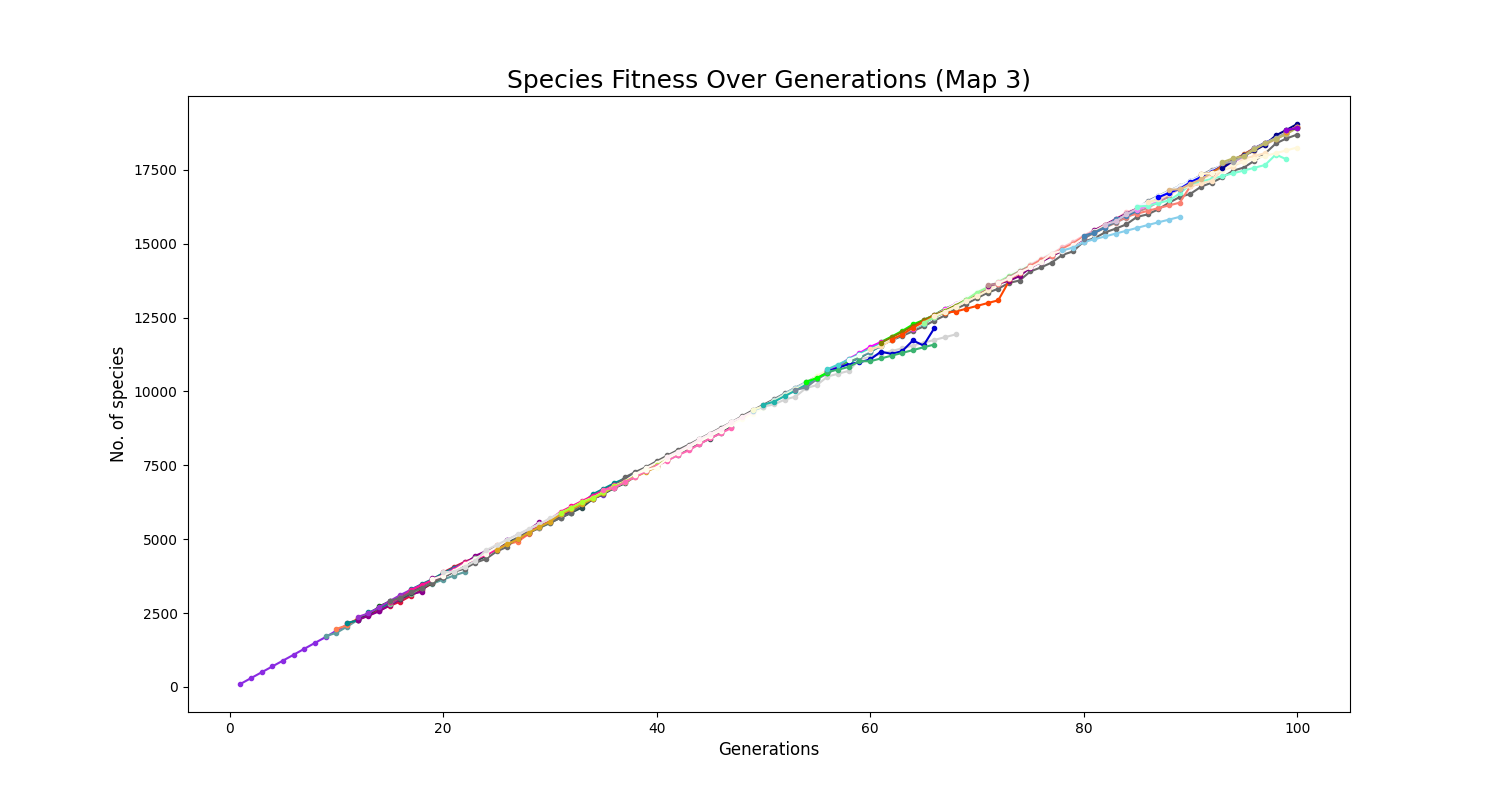}
    \caption{Species Fitness Over Generations Map 3}
    \label{fig:sfg3}
\end{figure}

\begin{figure}[ht]
    \centering
    \includegraphics[width=\linewidth]{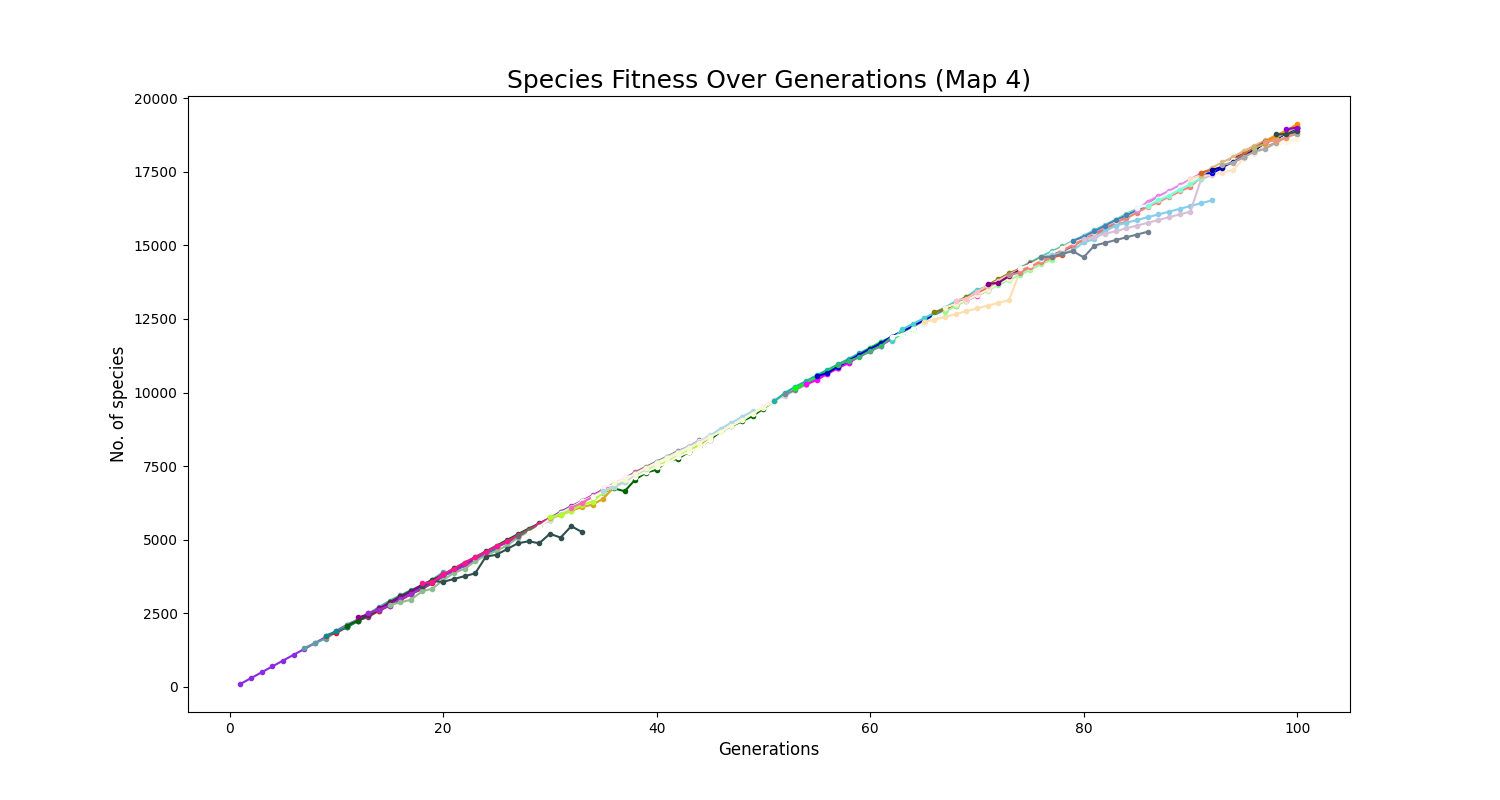}
    \caption{Species Fitness Over Generations Map 4}
    \label{fig:sfg4}
\end{figure}


\begin{figure}[ht]
    \centering
    \includegraphics[width=\linewidth]{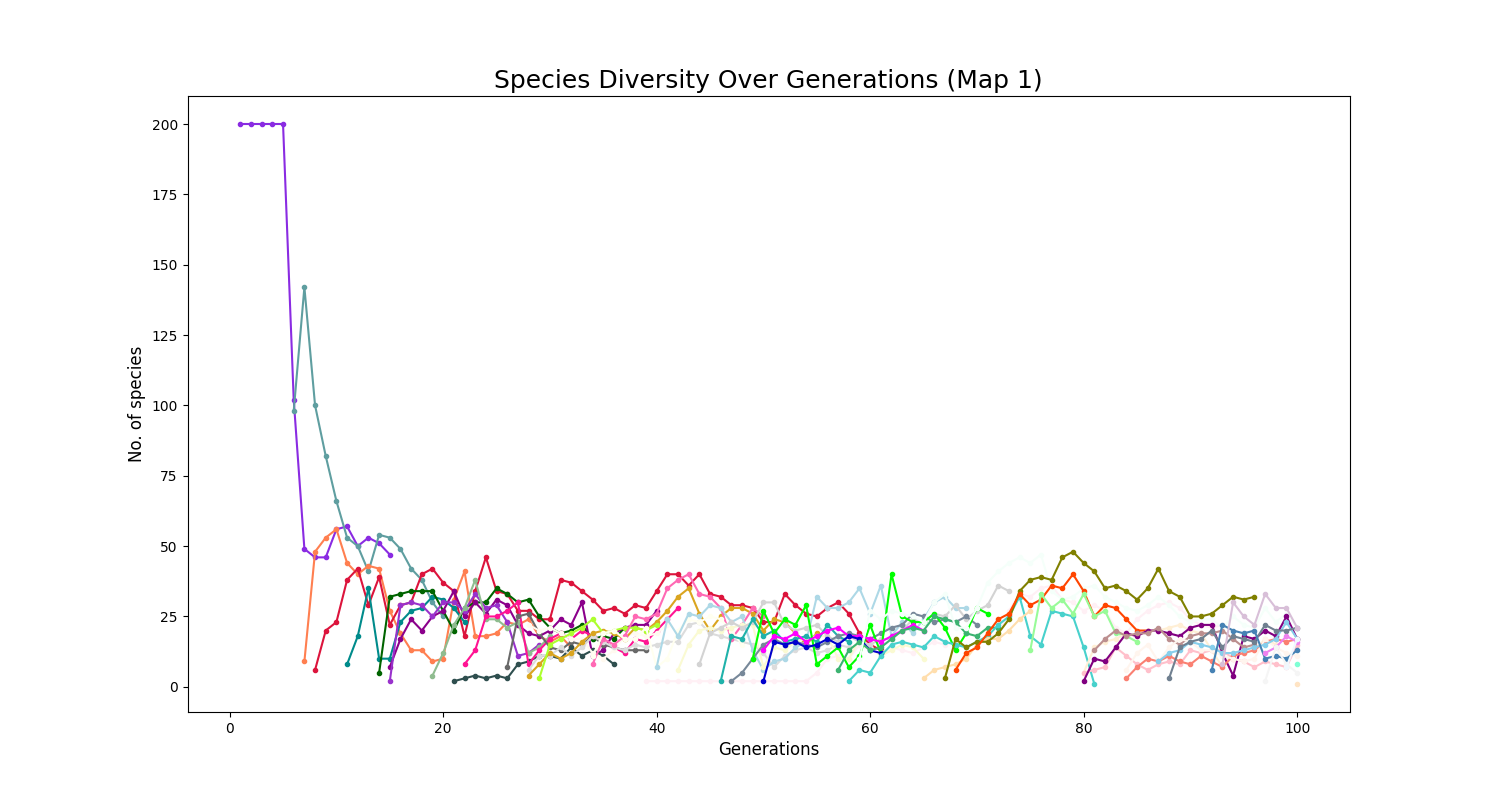}
    \caption{Species Diversity Over Generations Map 1}
    \label{fig:sdg1}
\end{figure}

\begin{figure}[ht]
    \centering
    \includegraphics[width=\linewidth]{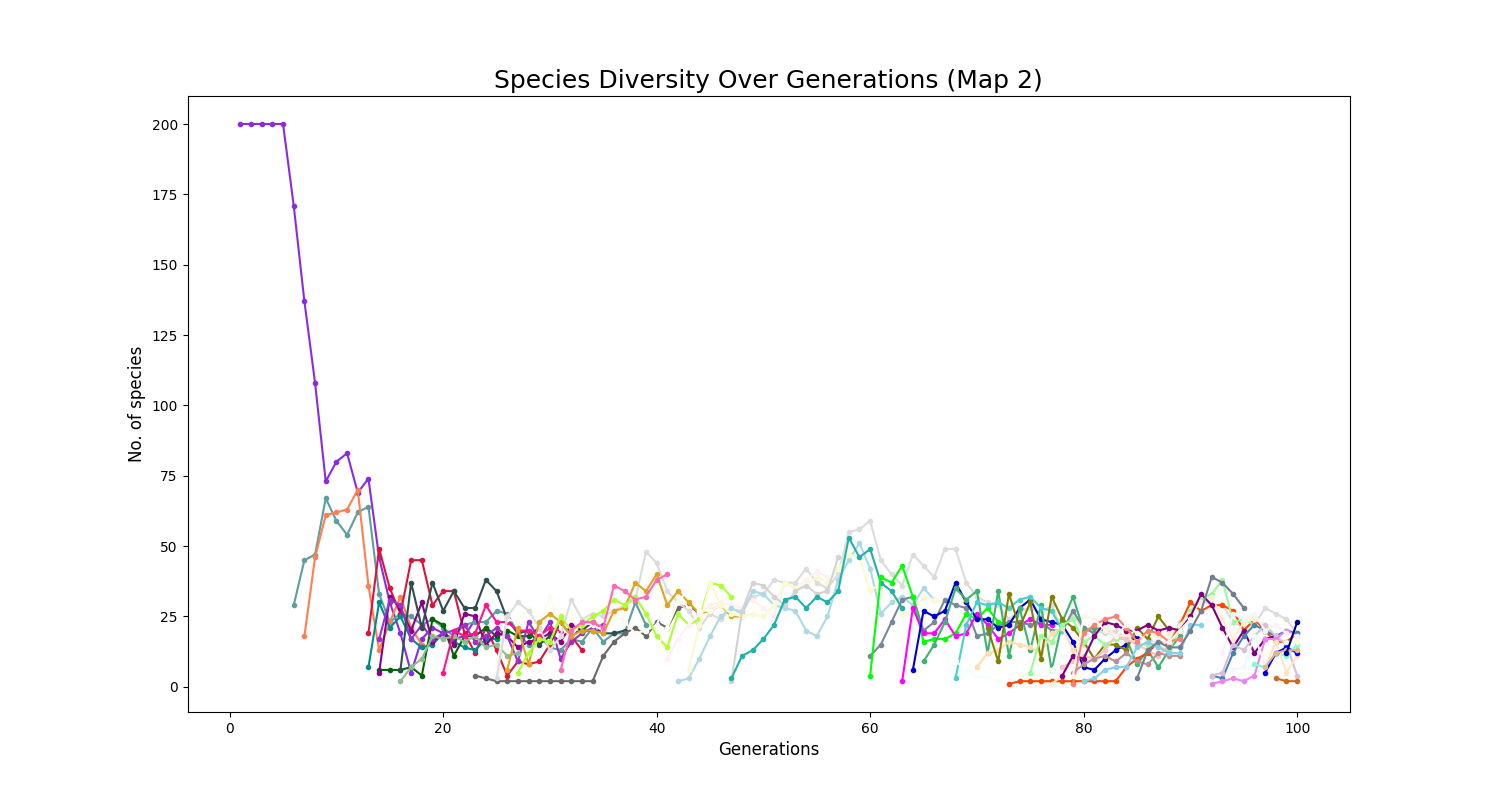}
    \caption{Species Diversity Over Generations Map 2}
    \label{fig:sdg2}
\end{figure}

\begin{figure}[ht]
    \centering
    \includegraphics[width=\linewidth]{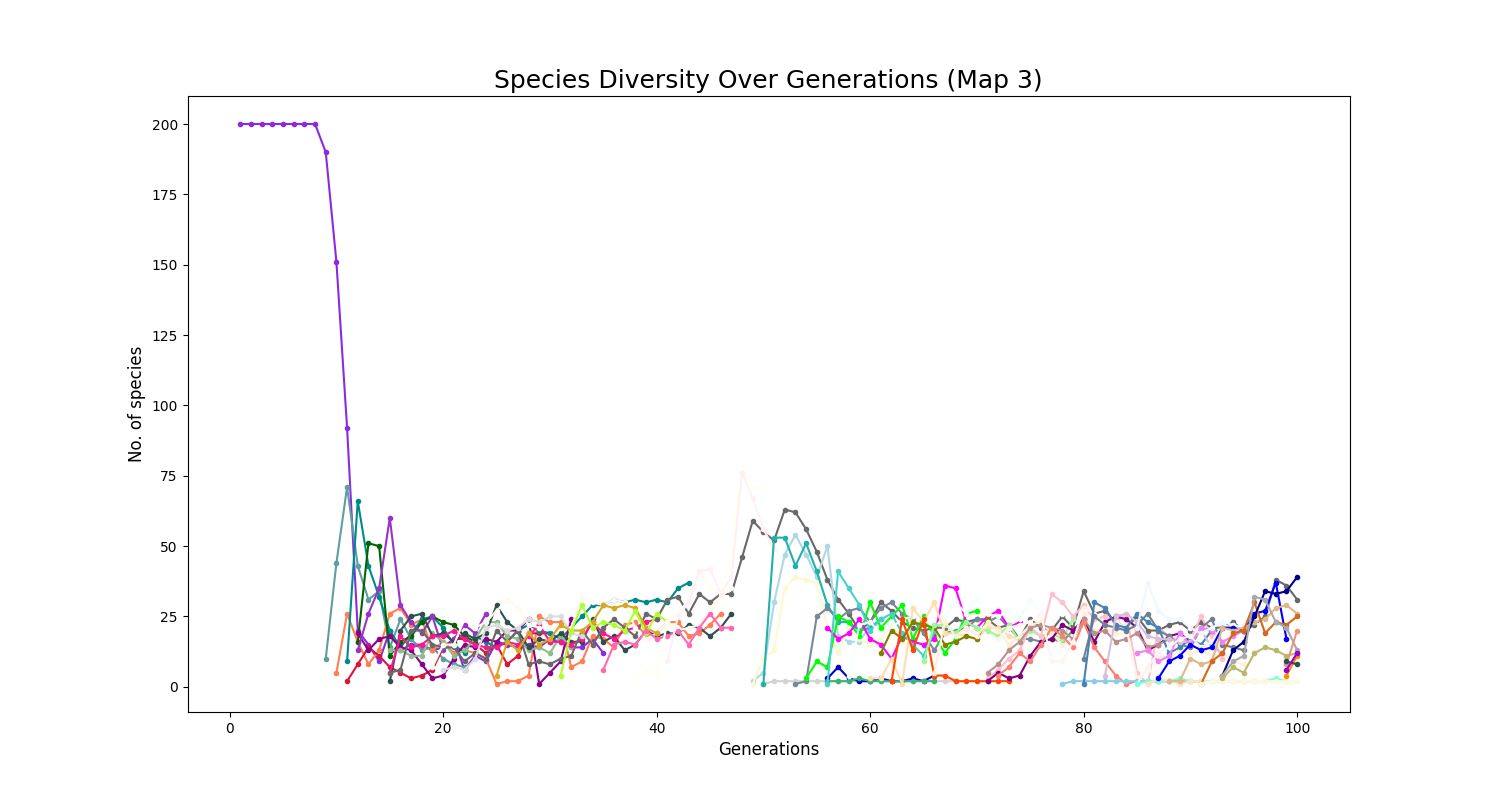}
    \caption{Species Diversity Over Generations Map 3}
    \label{fig:sdg3}
\end{figure}

\begin{figure}[ht]
    \centering
    \includegraphics[width=\linewidth]{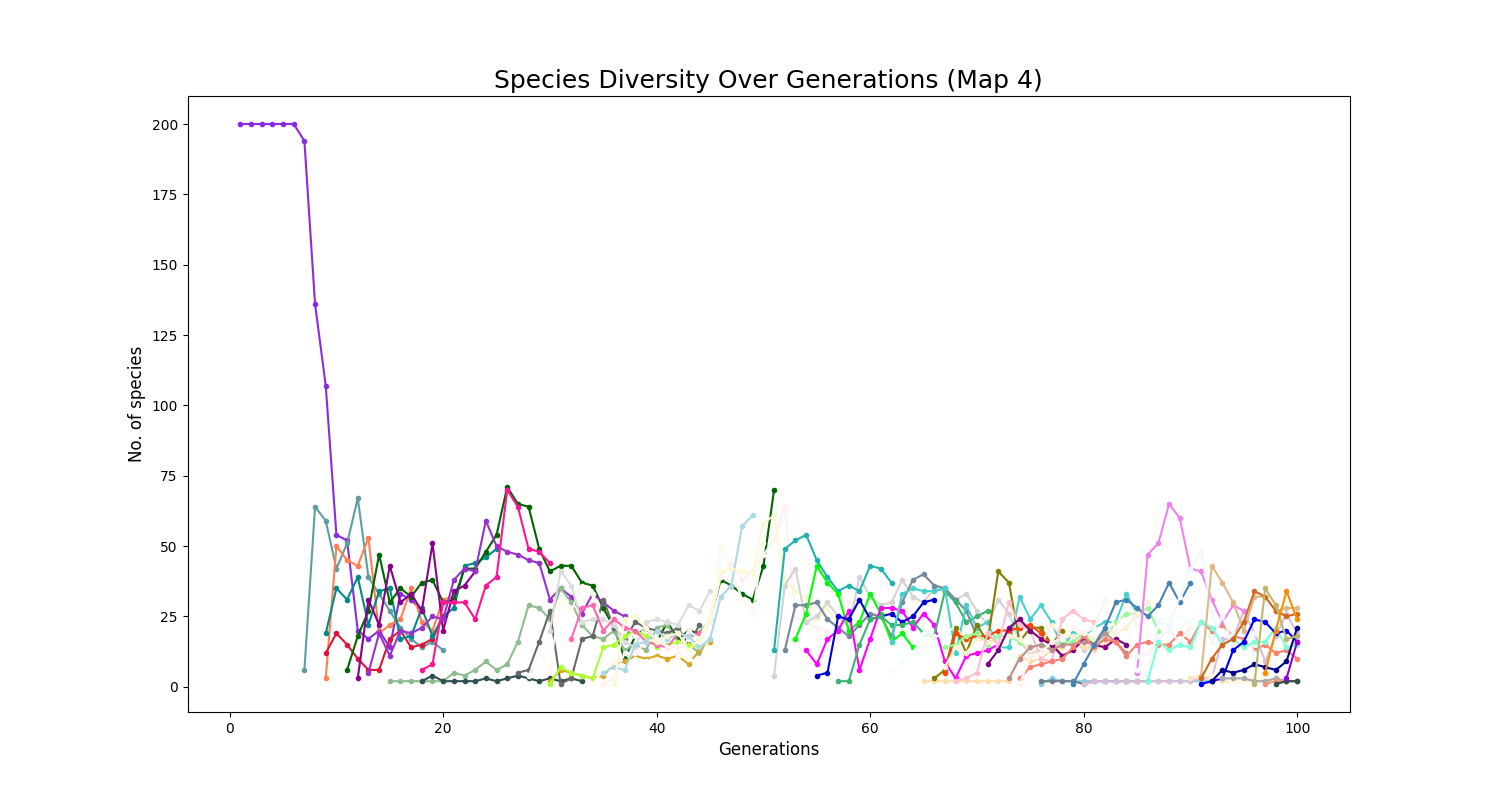}
    \caption{Species Diversity Over Generations Map 4}
    \label{fig:sdg4}
\end{figure}

\section{Conclusion}\label{conclu}

\end{document}